\pdfoutput=1
\documentclass{bmvc2k}
\usepackage{amsmath}

\newcommand{\suppref}[1]{supplementary~\cref{#1}}
\newcommand{\Suppref}[1]{Supplementary~\cref{#1}}
\usepackage{cleveref}
\usepackage{placeins}
\usepackage{stfloats}
\fnbelowfloat
\usepackage{amssymb}
\usepackage{bm}
\usepackage{verbatim}
\usepackage{algorithm}
\usepackage{algpseudocode}

\usepackage{booktabs} 
\usepackage{pgfplots}
\usepackage{url}
\makeatletter
\newcommand{\codefootnote}{\footnote{\ifbmv@review Code and data will be
released publicly upon publication.\else Project page, code and data:
\url{https://blobboards.github.io}\fi}}

\makeatother
\pgfplotsset{compat=1.18}
\newcommand{\fpscale}{1}

\newcommand{\bbDet}{80}
\newcommand{\bbAttTransMed}{4.5}
\newcommand{\bbAttRotMed}{0.58}

\newcommand{\bbRotMed}{0.46}

\newcommand{\bbTransMedLarge}{5.0}

\newcommand{\bbRotPninetyLarge}{0.98}

\newcommand{\bbTransMedMedium}{3.7}

\newcommand{\bbDetSmall}{78}
\newcommand{\bbTransMedSmall}{3.6}

\newcommand{\bbRotPninetySmall}{1.76}
\newcommand{\aprDet}{74}
\newcommand{\aprAttTransMed}{39.7}
\newcommand{\aprAttRotMed}{0.78}

\newcommand{\aprRotMed}{0.5}

\newcommand{\aprRotPninetyLarge}{1.22}

\newcommand{\aprDetSmall}{69}

\newcommand{\aprRotPninetySmall}{24.74}
\newcommand{\ucoDet}{58}
\newcommand{\ucoAttTransMed}{62.9}
\newcommand{\ucoAttRotMed}{12.55}

\newcommand{\ucoRotMed}{1.38}

\newcommand{\ucoDetSmall}{45}

\newcommand{\gainAprLarge}{3.3}
\newcommand{\gainAprLargePct}{70}
\newcommand{\gainAprMedium}{6.9}

\newcommand{\gainAprSmall}{9.1}
\newcommand{\gainAprSmallPct}{89}

\newcommand{\bbDetOccTen}{0.8}

\newcommand{\bbTransMedOccTen}{3.9}
\newcommand{\bbRotMedOccTen}{1.1}
\newcommand{\aprDetOccTen}{0.11}

\newcommand{\aprTransMedOccTen}{33.0}

\newcommand{\ucoDetOccTen}{0.23}

\newcommand{\ucoTransMedOccTen}{91.2}
\newcommand{\ucoRotMedOccTen}{18.79}

\newcommand{\bbDetPctOccFifty}{69}
\newcommand{\bbTransMedOccFifty}{3.9}
\newcommand{\bbRotMedOccFifty}{1.05}

\newcommand{\bbTransMedOccMin}{3.6}
\newcommand{\bbTransMedOccMax}{4.0}

\newcommand{\bbTimeMeanS}{2.03}
\newcommand{\bbTimeStdS}{2.18}

\newcommand{\aprTimeMeanS}{0.11}
\newcommand{\aprTimeStdS}{0.01}

\newcommand{\ucoTimeMeanS}{0.04}
\newcommand{\ucoTimeStdS}{0.01}

\newcommand{\bbDetPoses}{1346}

\newcommand{\bbMinSpacingMM}{47.8}

\newcommand{\bbHeFrames}{41}
\newcommand{\bbHeRecovered}{37}
\newcommand{\bbHeMarkers}{30}
\newcommand{\bbHeResidTransMM}{0.68}
\newcommand{\bbHeResidRotDeg}{0.27}

\newcommand{\ucoHeFrames}{15}

\newcommand{\flipApr}{124}
\newcommand{\flipUco}{152}
\newcommand{\flipBBraw}{45}
\newcommand{\flipBBlrt}{33}
\newcommand{\flipBBcorrected}{12}

\newcommand{\vtheta}{\bm{\theta}}
\newcommand{\va}{\bm{a}}

\graphicspath{{images}}

\makeatletter
\let\bmv@origaddauthor\addauthor
\def\addauthor#1#2#3{%
  \bmv@origaddauthor{#1}{#2}{#3}%
  \def\bmv@tmpmail{#2}%
  \expandafter\xdef\csname bmv@hasmail\the\bmv@nauthors\endcsname
    {\ifx\bmv@tmpmail\@empty 0\else 1\fi}}
\def\bmv@RenderInst#1{%
  \begin{minipage}[t]{0.28\textwidth}
    \sffamily
    \begin{raggedright}#1\end{raggedright}
  \end{minipage}}
\newcount\bmv@authcol
\newcount\bmv@gridauthors
\AtBeginDocument{\bmv@gridauthors=\bmv@nauthors \advance\bmv@gridauthors-1}
\def\bmv@authorcell#1{%
  \bmv@usebox{authname#1}%
  \ifnum\csname bmv@hasmail#1\endcsname=1
    \\ \bmv@usebox{authmail#1}%
  \fi}
\def\bmv@RenderAuthInstTwoColumn{%
  \begin{minipage}[t]{0.70\textwidth}
    \raggedright
    \bmv@authcol=0
    \def\bmv@action##1{%
      \ifnum##1>\bmv@gridauthors\else
        \begin{minipage}[t]{0.5\linewidth}%
          \raggedright
          \bmv@authorcell{##1}%
        \end{minipage}%
        \global\advance\bmv@authcol1
        \ifnum\bmv@authcol=2
          \global\bmv@authcol=0
          \ifnum##1<\bmv@gridauthors \\[3pt] \fi
        \fi
      \fi}%
    \def\bmv@between{}%
    \expandafter\bmv@maplistaux\bmv@auths\bmv@endstop
  \end{minipage}\hfill
  \begin{minipage}[t]{0.28\textwidth}
    \raggedright
    \def\bmv@action##1{\bmv@unhbox{inst##1}}%
    \def\bmv@between{\\[4pt]}%
    \expandafter\bmv@maplistaux\bmv@insts\bmv@endstop
    \\[\dimexpr2\baselineskip+3pt\relax]
    \bmv@authorcell{\the\bmv@nauthors}%
  \end{minipage}}
\makeatother

\title{BlobBoards: {\large Robust Markers for Accurate Pose\codefootnote}}

\addauthor{James Pritts}{jpr@informatik.uni-kiel.de}{1}
\addauthor{Till Sittart}{till.sittart@till-s.de}{1}
\addauthor{Hendrik Sauer}{hendrik-sauer@mailbox.org}{1}
\addauthor{Silja Jan\ss{}en}{sja@informatik.uni-kiel.de}{1}
\addauthor{Felix Seegr\"aber}{fse@informatik.uni-kiel.de}{1}
\addauthor{David Nakath}{dna@cs.uni-kiel.de}{1}
\addauthor{Kevin K\"oser}{kk@informatik.uni-kiel.de}{1}

\addinstitution{
 Marine Data Science\\
 Kiel University
}

\runninghead{Pritts \bmvaEtAl}{BlobBoards}

\def\etal{\emph{et al}\bmvaOneDot}

\begin{document}
\setlength{\emergencystretch}{3pt}

\maketitle

\begin{abstract}
	We propose BlobBoards, a fiducial marker system comprising a dense,
	multi-scale field of Gaussian blobs and a feature-based pipeline for
	joint detection, identification, and pose estimation. Each board is
	registered from hundreds of blob features whose dense
	spatial coverage constrains pose, while multiple scales preserve
	detectability across large changes in focal length, distance, and
	obliquity. Learned local descriptors are matched to the reference
	pattern and spatially verified, so the correspondences determine pose
	and certify identity. Against motion-capture ground truth, BlobBoards
	achieve median translation errors of
	$\bbTransMedSmall$--$\bbTransMedLarge$\,mm, reducing AprilTag's median
	translation error by $\gainAprSmallPct\%$ on small boards and
	$\gainAprLargePct\%$ on large ones. They also produce far fewer
	large-rotation failures than state-of-the-art tag systems. 
	BlobBoards achieve the highest detection rate, $\bbDet\%$ versus
	$\aprDet\%$ for AprilTag and $\ucoDet\%$ for ArUco, with the largest
	margin on the smallest markers. 
	Under $50\%$ occlusion, they still detect $\bbDetPctOccFifty\%$ of boards
	with essentially unchanged median translation error, while AprilTag and
	ArUco detect none. In experiments BlobBoards give state-of-the-art detection rate, pose accuracy and occlusion robustness.
\end{abstract}

\begin{figure}[!b]
	\centering
	\includegraphics[width=1.\linewidth]{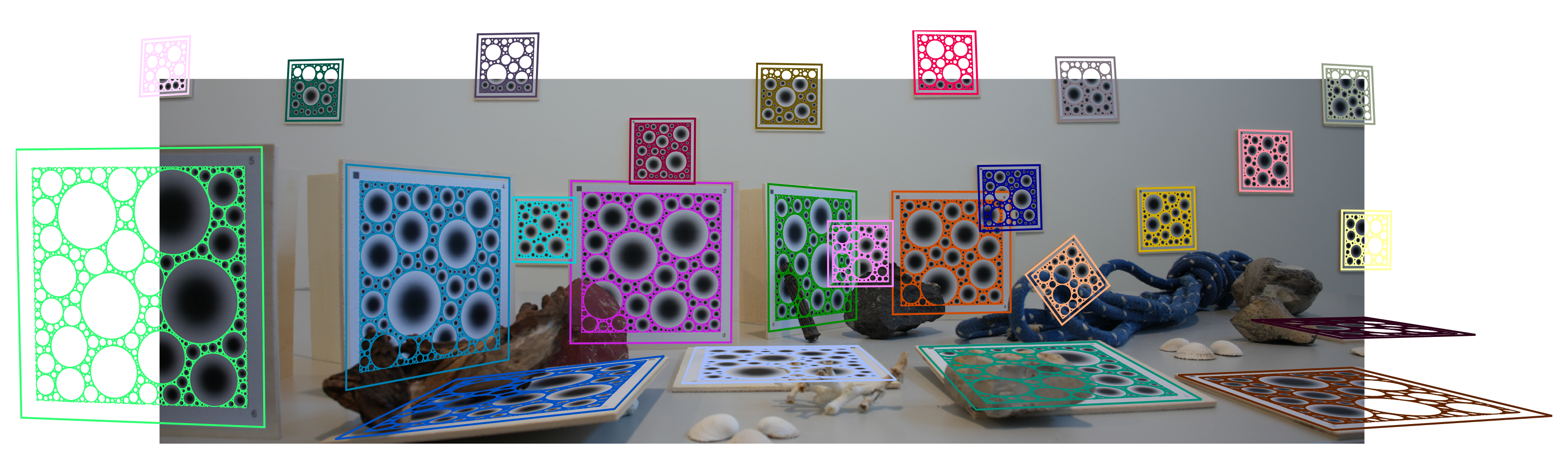}
	\caption{All $25$ BlobBoards in the image are detected, identified, and
		registered against a $50$-board gallery, so $25$ of its entries are
		absent confusers, and identification is still perfect. Colour
		distinguishes detections; the overlaid pattern shows registration
		accuracy across large variations in scale, distance, and obliquity.}
	\label{fig:teaser}
\end{figure}

\section{Introduction}
\label{sec:intro}
Fiducial markers are a foundational tool in robotics and computer
vision~\cite{kalaitzakis2021fiducial,kalaitzakis2020experimental},
providing reliable references for camera pose estimation and instance
identification in applications including robotic manipulation, SLAM,
augmented reality, multi-camera calibration, and autonomous navigation. A
useful marker should be accurately localizable, robust to viewpoint
and partial visibility, and uniquely identifiable so that multiple
markers can coexist in a scene.

Two design families dominate, and both designs inherently limit
keypoint density. Grid targets such as the checkerboard localize
saddle points to sub-pixel
accuracy~\cite{zhang2000flexible,Hartley00}, but each saddle needs
uniform regions of alternating black and white, and the repeated grid
carries no local identity: under partial visibility the correspondence
between observed corners and board coordinates becomes
ambiguous. Coded fiducials solve identity, but an independently
identifiable marker of the ARToolKit~\cite{kato1999marker},
ArUco~\cite{garrido2014automatic,romero2018speeded} or
AprilTag~\cite{olson2011apriltag,wang2016apriltag,krogius2019flexible}
family is a single high-contrast quadrilateral contributing exactly
four corner measurements regardless of its
size~\cite{kalaitzakis2020experimental,kalaitzakis2021fiducial}. Measurement
density per unit marker area therefore \emph{falls} as the marker
grows.  Occlusion of the segmented boundary disables the marker
outright. Multi-marker calibration boards
(AprilGrid~\cite{furgale2013unified},
ChArUco~\cite{garrido2014automatic,bradski2000opencv}), like the
self-identifying checkerboard variants
CALTag~\cite{atcheson2010caltag} and deltille
grids~\cite{ha2017deltille}, supply more measurements over a fixed
board area by tiling the target with markers. In that case, the target
is treated as a single rigid object and markers are used to place
joint constraints on the camera.

BlobBoards are designed to maximize the density of localizable
features. A board is a randomly generated field of Gaussian blobs
distributed densely over its area and across multiple scales
(\cref{fig:teaser}). Measurement density is a property of the pattern
rather than of the board size. The markers used in our evaluation
carry between $213$ and $1027$ blobs each over $4$ to $12$\,cm
footprints. High spatial density yields a well-constrained pose
estimate from many distributed correspondences, while the scale
distribution ensures that some subset of blobs remains at a feasible
detection scale as focal length, distance and viewpoint change.

A robust affine adaptation localizes each blob to sub-pixel accuracy. Each adapted
blob is canonicalized and described by the texture of its surrounding
region, whose random spatial configuration is unique to the board.
Tentative descriptor matches are spatially verified by registering the
board to its image in a RANSAC estimator. Pose estimation therefore
identifies and registers the board simultaneously.

Other markers have also left the
square. RUNE-Tag~\cite{bergamasco2011runetag} arranges dots on
concentric rings and CCTag~\cite{calvet2016cctag} localizes concentric
contours to sub-pixel precision; both raise per-marker accuracy but
keep a fixed, designed feature set. Closer still are codeless patterns
recognized from the configuration of their own features: random dot
markers~\cite{uchiyama2011random} identify a marker from the local
arrangement of scattered points, and Li~\etal~\cite{li2013multiple}
build a calibration pattern from descriptor-matched
features. BlobBoards keep that codeless principle but replace point
dots with affine-covariant Gaussian blobs at many scales, so that each
feature is independently localizable under viewpoint change and the
feature count grows with printed area.

Evaluated against motion-capture ground truth, BlobBoard's dense
feature coverage substantially improves absolute-pose accuracy over
AprilTag and ArUco, with the advantage increasing as markers become
smaller and more oblique. On the same captures BlobBoards also detect
the largest fraction of visible boards, and this detection lead is
likewise largest on the smallest markers, so BlobBoards improve on both
detection and accuracy at once. Under occlusion, BlobBoard degrades gracefully because
occluded regions remove local features rather than disabling the marker.

\section{BlobBoards}
\label{sec:blob_boards}

\begin{figure}[t]
	\centering
	\begin{minipage}{0.48\textwidth}
		\centering
		{\small maximum diameter fraction}\\[2pt]
		\begin{minipage}{0.32\textwidth}
			\includegraphics[width=\textwidth]{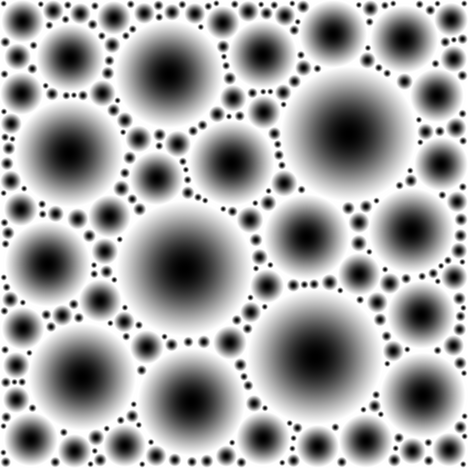}
		\end{minipage}
		\hfill
		\begin{minipage}{0.32\textwidth}
			\includegraphics[width=\textwidth]{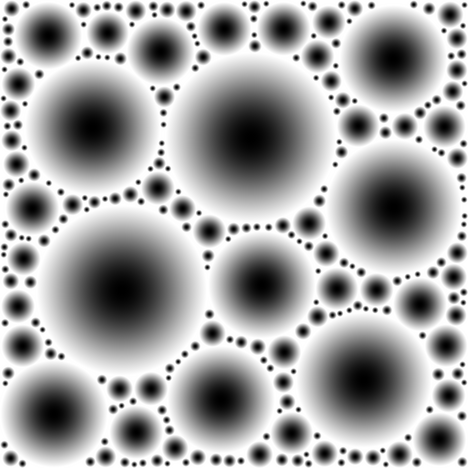}
		\end{minipage}
		\hfill
		\begin{minipage}{0.32\textwidth}
			\includegraphics[width=\textwidth]{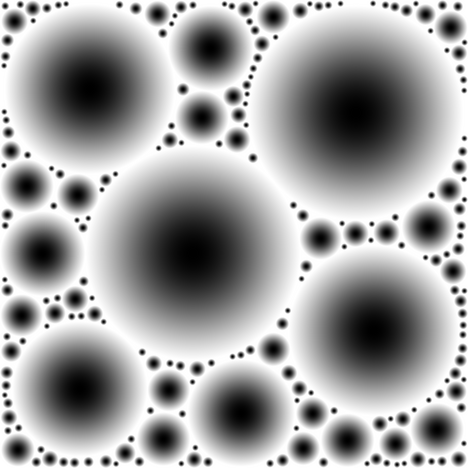}
		\end{minipage}
	\end{minipage}
	\hfill
	\begin{minipage}{0.48\textwidth}
		\centering
		{\small minimum blob scale}\\[2pt]
		\begin{minipage}{0.32\textwidth}
			\includegraphics[width=\textwidth]{arxiv/images/blob_variations_squared/blob_pattern_259_e8a21_co1.5_a1.2_df0.4_s0.2.png}
		\end{minipage}
		\hfill
		\begin{minipage}{0.32\textwidth}
			\includegraphics[width=\textwidth]{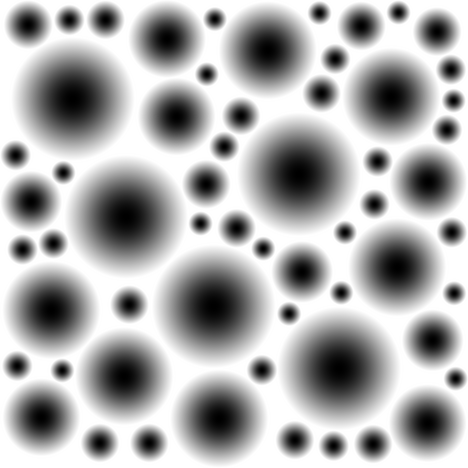}
		\end{minipage}
		\hfill
		\begin{minipage}{0.32\textwidth}
			\includegraphics[width=\textwidth]{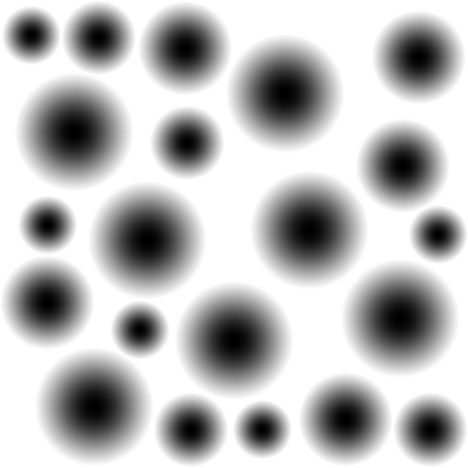}
		\end{minipage}
	\end{minipage}

	\vspace{6pt}

	\begin{minipage}{0.48\textwidth}
		\centering
		{\small support cutoff $c$}\\[2pt]
		\begin{minipage}{0.32\textwidth}
			\includegraphics[width=\textwidth]{arxiv/images/blob_variations_squared/blob_pattern_259_e8a21_co1.5_a1.2_df0.4_s0.2.png}
		\end{minipage}
		\hfill
		\begin{minipage}{0.32\textwidth}
			\includegraphics[width=\textwidth]{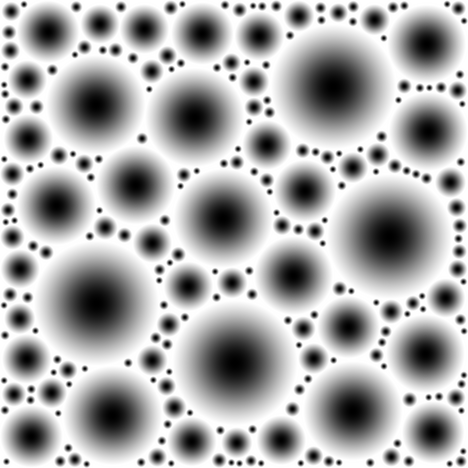}
		\end{minipage}
		\hfill
		\begin{minipage}{0.32\textwidth}
			\includegraphics[width=\textwidth]{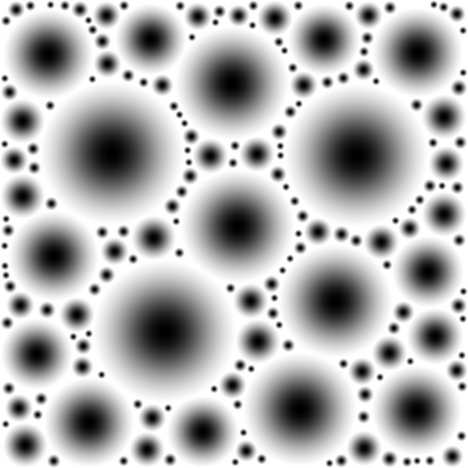}
		\end{minipage}
	\end{minipage}
	\hfill
	\begin{minipage}{0.48\textwidth}
		\centering
		{\small scale-sampling distribution}\\[2pt]
		\begin{minipage}{0.32\textwidth}
			\includegraphics[width=\textwidth]{arxiv/images/blob_variations_squared/blob_pattern_259_e8a21_co1.5_a1.2_df0.4_s0.2.png}
		\end{minipage}
		\hfill
		\begin{minipage}{0.32\textwidth}
			\includegraphics[width=\textwidth]{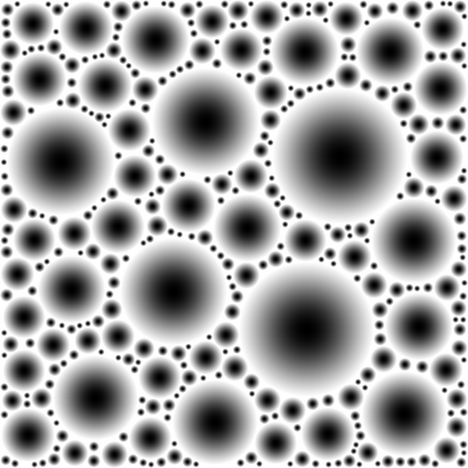}
		\end{minipage}
		\hfill
		\begin{minipage}{0.32\textwidth}
			\includegraphics[width=\textwidth]{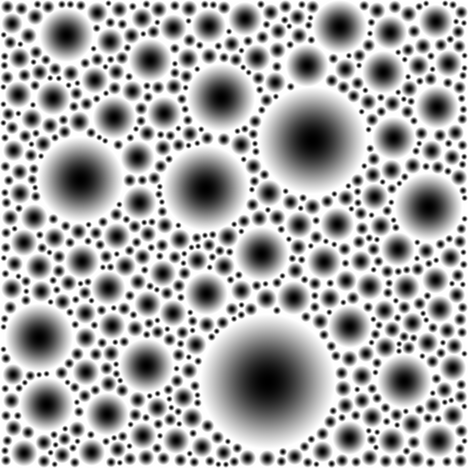}
		\end{minipage}
	\end{minipage}
	\caption{Effect of changing BlobBoard generation parameters.}
	\label{fig:params}
\end{figure}

A BlobBoard is a dense random packing of dark Gaussian blobs at
different positions and scales. Each blob of scale $\sigma_0$ is
truncated at radius $c\sigma_0$ and shifted and normalized to meet the
white background continuously:
\begin{equation}
	f(\mathbf{x}) = \left(1 -
	\exp\!\left(-\|\mathbf{x}\|^2/2\sigma_0^2\right)\right) \big/
	\left(1 - \exp\!\left(-c^2/2\right)\right), \quad \|\mathbf{x}\|
	\le c\sigma_0, \quad \mathbf{x}\in \mathbb{R}^2
	\label{eq:blobmodel}
\end{equation}
The compact support allows blobs to be packed densely without overlap,
while varying $\sigma_0$ makes the pattern intrinsically multi-scale.

Up to the additive white background, an ideal blob is a negative isotropic
Gaussian
\[
	I(\mathbf{x}) = -A\,\exp\!\left(-\|\mathbf{x}\|^2/2\sigma_0^2\right)
\]
of contrast~$A$. SIFT localizes features as scale-space extrema of the
scale-nor\-mal\-ized Lapla\-cian-of-Gauss\-ian
\[
	R(\mathbf{x},\sigma) = \sigma^2\,\nabla^2(G_\sigma * I)(\mathbf{x}),
\]
where $G_\sigma$ is the isotropic Gaussian kernel of standard deviation
$\sigma$ and $*$ denotes convolution; the Difference-of-Gaussians pyramid
approximates this operator~\cite{lowe2004distinctive,lindeberg1998feature}. Since the
Gaussian family is closed under convolution, the response at the blob centre
is
\begin{equation}
	R(\mathbf{0},\sigma)
	\;=\; -\,2A\,\sigma_0^2\,\sigma^2 \big/ (\sigma_0^2+\sigma^2)^2,
	\qquad
	\bigl|R(\mathbf{0},\sigma)\bigr|\ \text{maximal at}\ \sigma=\sigma_0.
	\label{eq:logresponse}
\end{equation}
The maximum response is $A/2$, independent of $\sigma_0$. An ideal Gaussian
therefore has two useful properties for a multi-scale target. First, it is
scale-covariant: the scale-space extremum occurs at the blob centre and its
true scale. Second, its peak response is scale-independent: fine and coarse
blobs of equal contrast respond equally strongly. Truncating the Gaussian
gives it the compact support required to pack these accurately localized
features at high spatial density, which is central to pose accuracy.

The finite-support profile of \cref{eq:blobmodel} preserves both properties
but introduces a predictable scale bias. Truncation alone perturbs the matched
scale only slightly; the dominant effect is the removal of the pedestal
$p=e^{-c^2/2}$ that makes the blob meet the background continuously.
Carrying this through the centre response gives an extremum at
$\beta(c)\,\sigma_0$. We call $\beta(c)$ the \emph{pedestal scale bias}: it
depends only on the support cutoff $c$ and not on $\sigma_0$, so the response
stays scale-covariant with a scale-independent peak, and each detected scale
is rescaled by the \emph{pedestal correction factor} $1/\beta(c)$ to recover
$\sigma_0$ before the affine adaptation of \cref{sec:adapt}. \Suppref{sec:supp:derivations} gives the closed form of the truncated
response (\cref{eq:supp:truncresponse}) and of~$\beta$.

\noindent\textbf{Board generation.}~ We generate each board by greedily
packing blobs from a discrete set of scales while keeping their supports
disjoint. Blob $j$ is defined by $(\mathbf{c}_j,\sigma_j)$, where
$\mathbf{c}_j$ is its centre in the metric coordinates of the printed pattern
and $\sigma_j$ its scale; its support is the disc of radius $c\sigma_j$
(\cref{eq:blobmodel}). Scales are processed coarse-to-fine, placing large
blobs first and using progressively smaller blobs to fill the remaining gaps.
Candidate centres are sampled uniformly over the board and accepted only if
their support does not overlap any placed blob. An R-tree accelerates overlap
queries, and a placement is abandoned once its fixed trial budget is
exhausted. The procedure is seeded for exact reproducibility, and is stated
in full as \suppref{alg:gen}.

Each accepted blob is then rasterised using \cref{eq:blobmodel} with
$\sigma_0=\sigma_j$. Because supports are disjoint, no compositing
rule is required. The generator outputs the printable image and the
reference geometry
$\mathcal{B}=\{(\mathbf{c}_j,\sigma_j)\}_{j=1}^{N}$,
which is used for registration. Random placement also makes the
neighbourhood of each blob distinctive; this surrounding texture,
rather than the Gaussian itself, provides the appearance cue used for
identification in \cref{sec:canonicalize_describe}.

Board generation is controlled by the minimum blob scale,
the maximum diameter fraction, the support cutoff $c$, and the
scale-sampling distribution, which trade off density, scale coverage, and
printability (\cref{fig:params}). Their values depend on working distance,
camera resolution, and print resolution; the settings used in our experiments
are given in \suppref{tab:supp:params}.

\section{The BlobBoard Pipeline}
\label{sec:pipeline}
\begin{figure*}[t]
	\centering
	\begin{minipage}{0.245\textwidth}
		\centering
		{\small (a) detect}\\[2pt]
		\includegraphics[width=\linewidth]{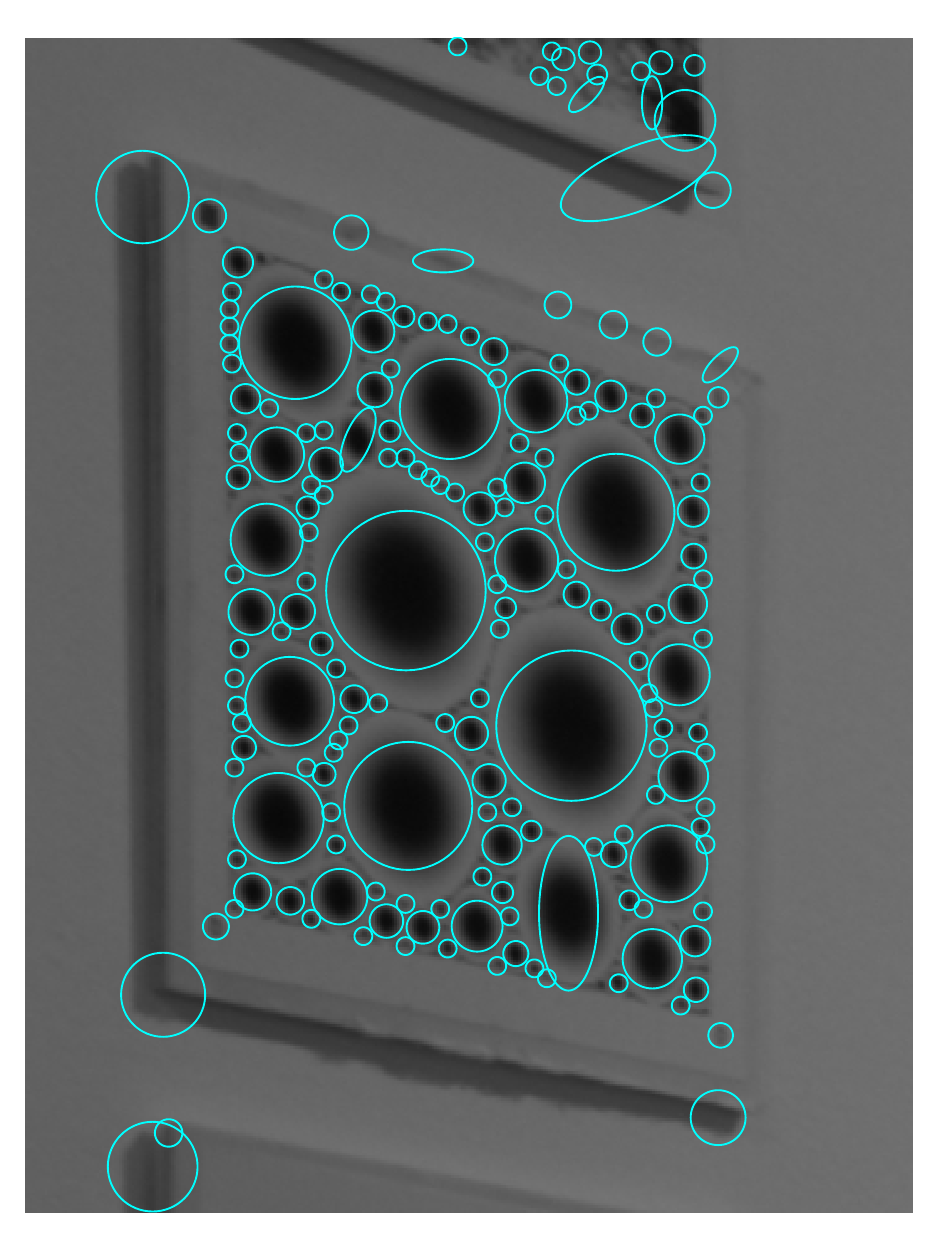}
	\end{minipage}\hfill
	\begin{minipage}{0.245\textwidth}
		\centering
		{\small (b) adapt}\\[2pt]
		\includegraphics[width=\linewidth]{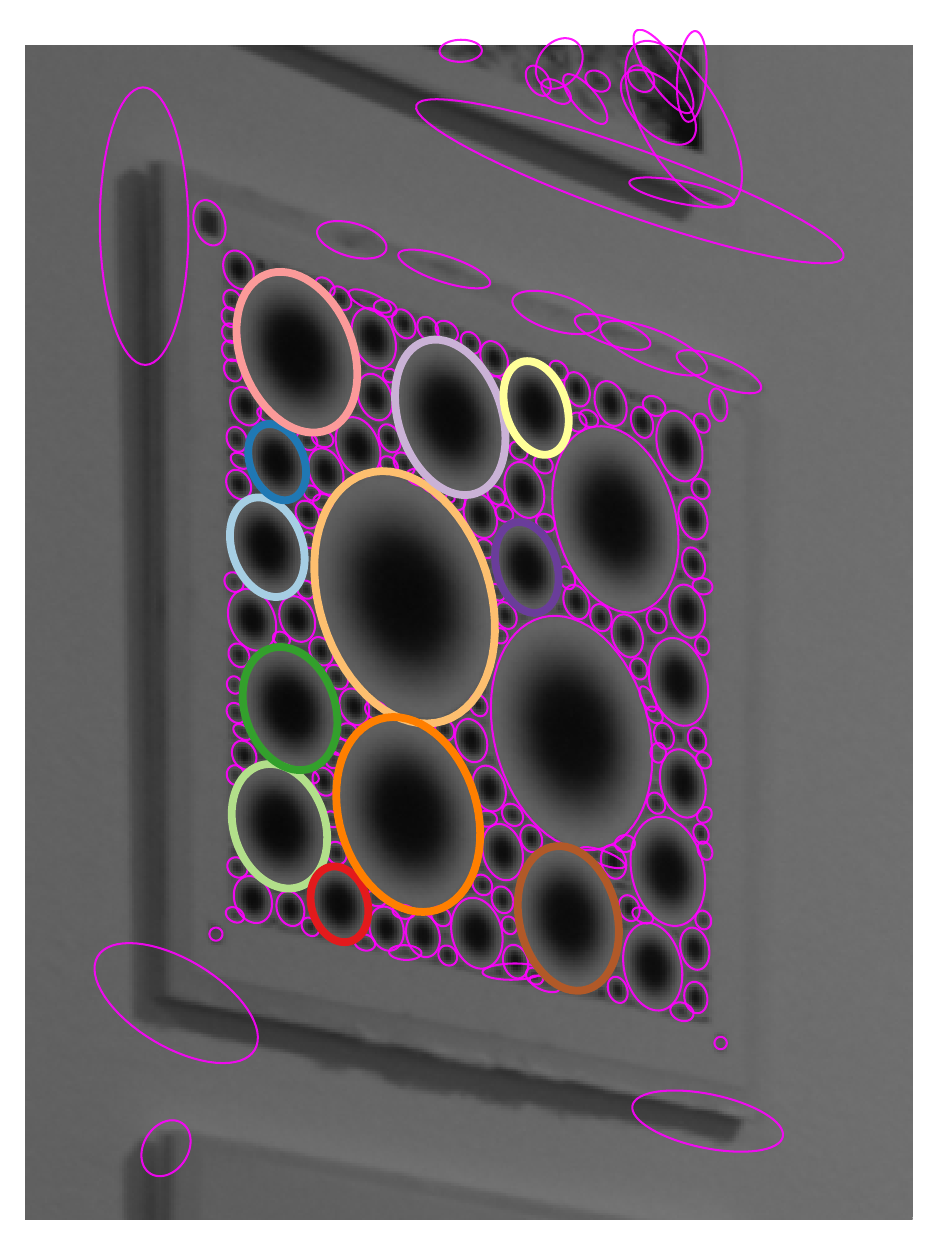}
	\end{minipage}\hfill
	\begin{minipage}{0.245\textwidth}
		\centering
		{\small (c) canonicalize}\\[2pt]
		\includegraphics[width=\linewidth]{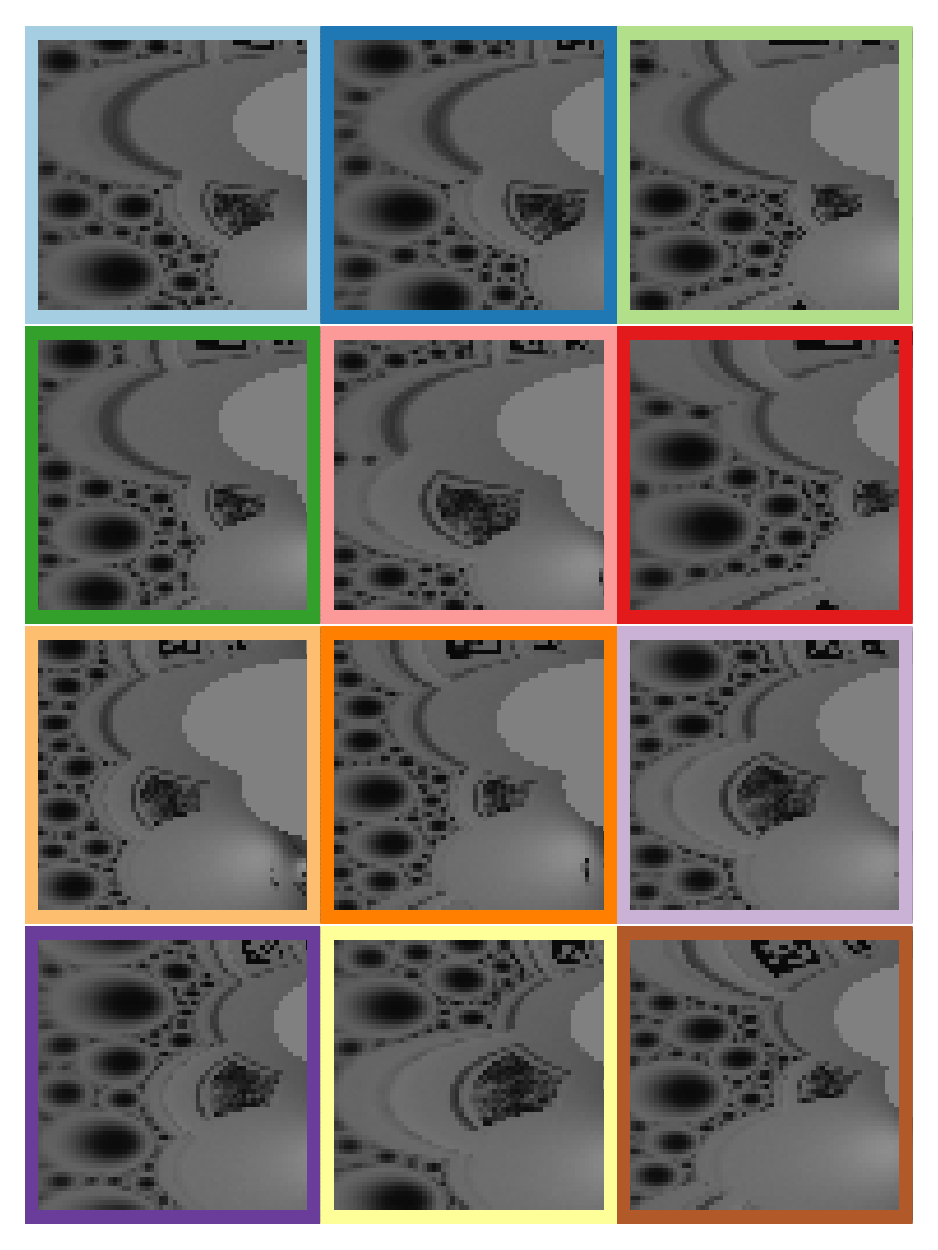}
	\end{minipage}\hfill
	\begin{minipage}{0.245\textwidth}
		\centering
		{\small (d) match \& verify}\\[2pt]
		\includegraphics[width=\linewidth]{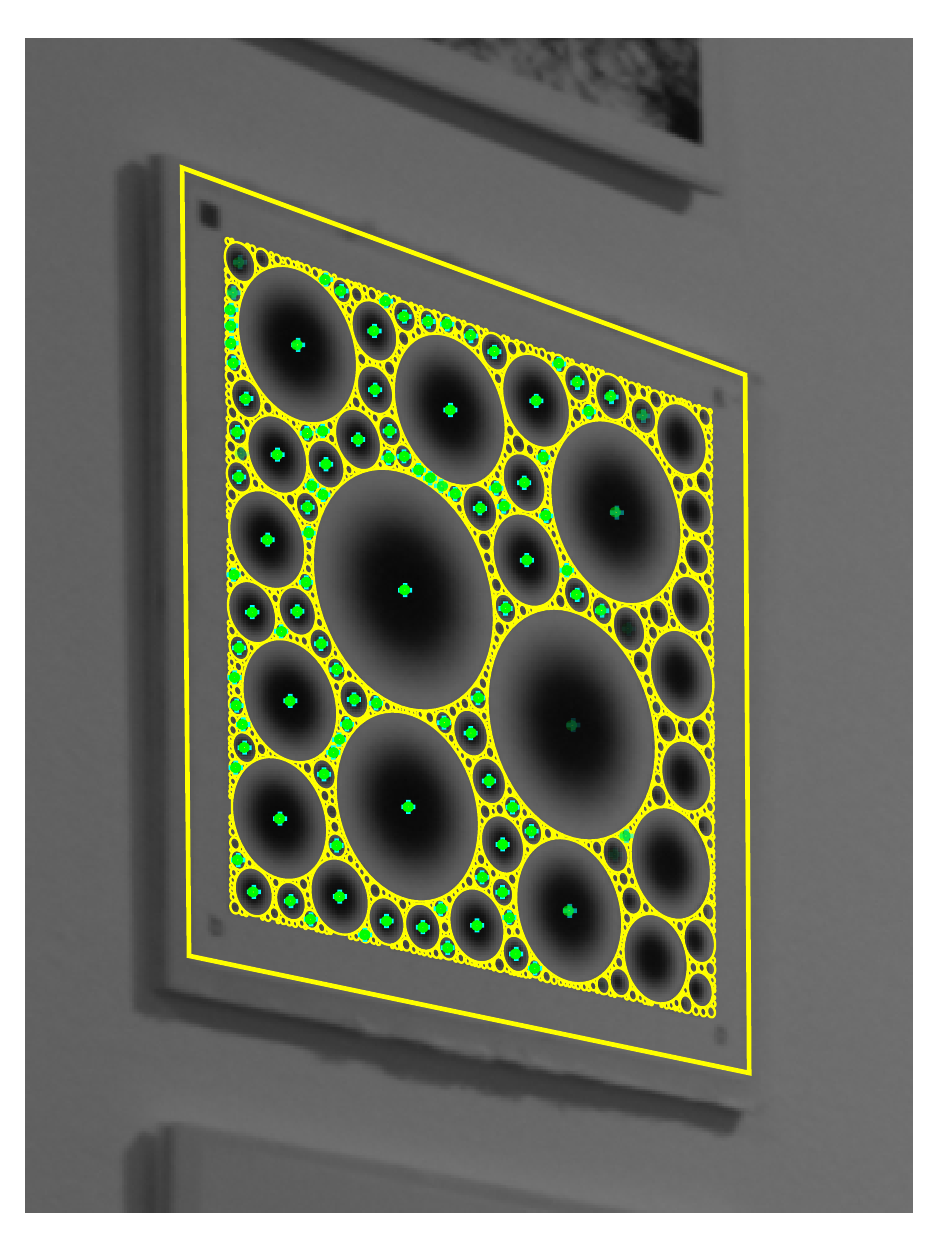}
	\end{minipage}
	\caption{Blob detection-to-pose pipeline on a detected board.
		\textbf{(a)}~Affine detections from the ASIFT scale space.
		\textbf{(b)}~IRLS adaptation to anisotropic ellipses, magenta except the
		twelve pose inliers, which are coloured individually.
		\textbf{(c)}~Those inliers canonicalized to $64{\times}64$ log-polar
		patches, each bordered in its ellipse colour from~(b).
		\textbf{(d)}~Descriptors matched and spatially verified: the pattern is
		projected under the recovered pose, with reference blob centres in
		green at opacity proportional to their geometric consistency with it.}
	\label{fig:pipeline}
\end{figure*}

BlobBoards detect, identify, and pose boards by formulating detection
as spatial verification~\cite{philbin2007object} over many local
features (\cref{fig:pipeline}). The pipeline has eight stages, grouped as
below: \textbf{(1)~detect} blob candidates in scale space and initialize an
elliptic frame; \textbf{(2)~adapt} that frame to the local image, recovering
the blob's characteristic scales and affine shape $\boldsymbol{\Sigma}$
(\cref{sec:detect}); \textbf{(3)~canonicalize} the surrounding patch by
whitening the ellipse and resampling it on a log-polar grid;
\textbf{(4)~describe} the patch with a learned 128-D embedding
(\cref{sec:canonicalize_describe}); \textbf{(5)~match} image descriptors
against a per-board reference gallery; \textbf{(6)~spatially verify} the
resulting correspondences with
LO-RANSAC~\cite{fischler1981random,chum2003locally,pritts2026scoring}, which
proposes and locally refines a geometric hypothesis; \textbf{(7)~certify} the
board's identity from the appearance of its verified inliers; and
\textbf{(8)~claim} the detections of each committed board, so that boards
competing for the same features are resolved in rounds
(\cref{sec:match_verify}). A likelihood-ratio test then resolves the two-fold
planar pose ambiguity~\cite{schweighofer2006robust}
(\cref{sec:orientation}). \Suppref{alg:pipeline} states the whole pipeline
in one place; the remainder of this section details each stage.

\subsection{Detect \& Adapt}
\label{sec:detect}
\label{sec:adapt}

\noindent\textbf{Detect.}~ We detect blob candidates with a GPU implementation of
ASIFT~\cite{morel2009asift,yu2011asift_ipol}: a Difference-of-Gaussians scale
space evaluated over a bank of $25$ synthetic affine warps --- an identity
warp plus rotations at two tilt levels spanning the target viewpoint range.
Extrema are localized to sub-pixel position $\boldsymbol{\mu}_0$ and
continuous scale $\sigma$, which the pedestal correction factor $1/\beta(c)$
maps back to the printed scale (\cref{sec:blob_boards}). An extremum is
isotropic in the warped image; pulled back through the inverse warp it is an
ellipse in the input image whose size is the detected characteristic scale
and whose aspect and orientation are the warp's. This ellipse is the seed
shape matrix $\boldsymbol{\Sigma}_0$. A physical blob may fire under several warps, so overlapping
ellipses are non-maximally suppressed by response. Detection returns the seed elliptic frame
$\mathcal{E}_0=(\boldsymbol{\mu}_0,\boldsymbol{\Sigma}_0)$ with its response
and polarity.

\noindent\textbf{Adapt.}~ The seed $\boldsymbol{\Sigma}_0$ thus
combines the characteristic scale of the detection with the shape of
the warp that produced it. It is an initial estimate of the imaged
blob's scale and shape.  Adaptation refines
$(\boldsymbol{\mu}_0,\boldsymbol{\Sigma}_0)$ against the local pixel
data. The refined shape matrix $\boldsymbol{\Sigma}\succ0$ encodes the
blob's two characteristic scales and its orientation, and its $c^2$
level set is the elliptic frame used for canonicalization
(\cref{sec:canonicalize_describe}). We model the local intensity as a
dark Gaussian on a bright background,
\begin{equation}
	I(\mathbf{x})
	\approx
	b - A \exp\!\left(
	-\tfrac{1}{2}(\mathbf{x}-\boldsymbol{\mu})^{\top}
	\boldsymbol{\Sigma}^{-1}
		(\mathbf{x}-\boldsymbol{\mu})
	\right),
	\qquad
	\mathbf{x} = (x,y)^{\top},
	\label{eq:affine:dark-blob}
\end{equation}
with $b$ estimated from the local maximum intensity in the fitting window,
$A>0$ the contrast and $\boldsymbol{\mu}$ the blob centre.

Fitting \cref{eq:affine:dark-blob} directly is nonlinear through the
exponential. Writing $f_i = b-I(\mathbf{x}_i)$ for the inverted intensity and
$\va_i=(1,\,x_i,\,y_i,\,x_i^2,\,2x_iy_i,\,y_i^2)^{\top}$, taking logarithms
over the samples with $f_i>0$ gives
\begin{equation}
	z_i \;\equiv\; \log f_i \;=\; \vtheta^{\top}\va_i,
	\label{eq:log-model}
\end{equation}
where $\vtheta\in\mathbb{R}^6$ collects the coefficients of the expanded
quadratic: the quadratic terms give the upper triangle of
$-\tfrac{1}{2}\boldsymbol{\Sigma}^{-1}$, the linear terms then give
$\boldsymbol{\Sigma}^{-1}\boldsymbol{\mu}$ and hence $\boldsymbol{\mu}$, and
the constant term $\log A-\tfrac12\boldsymbol{\mu}^{\top}
	\boldsymbol{\Sigma}^{-1}\boldsymbol{\mu}$ then gives $A$. Adaptation is
therefore a quadratic surface fit to the log-intensity profile.

The logarithm makes homoscedastic intensity noise heteroscedastic: to first
order, intensity noise of variance $\varsigma^2$ induces log-space variance
$\varsigma^2/f_i^2$. The corresponding generalized least squares (GLS) weight is therefore proportional to
$f_i^2$ --- equivalently, the residuals are whitened by $f_i$ --- giving
\begin{equation}
	L(\vtheta) \;\approx\; \sum_i \tilde
	w_i\,\bigl(\log f_i - \vtheta^{\top}\va_i\bigr)^2, \qquad \tilde
	w_i = w_i\,f_i^{\,2},
	\label{eq:whitened-obj}
\end{equation}
where $f_i^2$ is the deterministic GLS weight and $w_i$ is a robust
weight from a Student-$t$/uniform mixture whose inlier partition is
re-selected at every iteration by the same exact marginal-likelihood
sweep over the sorted residuals~\cite{pritts2026scoring} --- rejecting
neighbouring blobs, occluder edges and background clutter with no noise
scale to tune, since it is marginalized. Since $f_i^2=(\partial f/\partial z)^2$,
\cref{eq:whitened-obj} is the first-order intensity-space objective
expressed in linearized coordinates. It is solved by M-estimation~\cite{huber1981robust,bosse2016robust} using iteratively
reweighted least squares (IRLS) initialized from $\boldsymbol{\Sigma}_0$. We batch the solve on the GPU over all
detections.  Each valid detection returns the refined elliptic frame
$\mathcal{E} = (\boldsymbol{\mu}, \boldsymbol{\Sigma})$.

\subsection{Canonicalize \& Describe}
\label{sec:canonicalize_describe}

\noindent\textbf{Canonicalize.}~ \label{sec:canonicalize}
To first order, perspective imaging maps the neighbourhood of a planar blob by
a local affinity. The adapted shape matrix $\boldsymbol{\Sigma}$ captures this
deformation, and the whitening transform
\[
	\mathbf{x}' = \boldsymbol{\Sigma}^{-1/2}(\mathbf{x}-\boldsymbol{\mu})
\]
maps the blob's elliptic frame to the circle
$\|\mathbf{x}'\|=c$~\cite{baumberg2000reliable}, removing local affine
deformation up to an in-plane rotation. A log-polar
resampling~\cite{esteves2017polar,ebel2019beyond} then
maps this residual rotation to translation along the angular axis.

The Gaussian blob is not discriminative, since every blob
has the same radial profile. The unique signal lies in the random
configuration of neighboring blobs. We sample an annulus outside
the blob in whitened coordinates, where radii are in units of the blob's
scale: the inner radius is the support cutoff $c$ and the outer radius the
hyperparameter $c_{\mathrm{out}}$, and we resample
$r=\|\mathbf{x}'\|\in[c, c_{\mathrm{out}}]$ onto a fixed $64\times64$
log-polar grid. Whitening and log-polar resampling
are combined into a single per-feature warp, yielding an affine-normalized
representation of the surrounding board texture.

\begin{figure}[t]
	\centering
	\includegraphics[width=\linewidth]{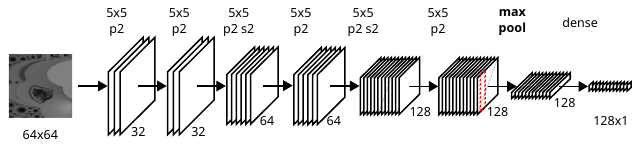}
	\caption{Descriptor architecture, adapted from
		HardNet~\cite{mishchuk2017working}: $64\times64$ log-polar input,
		$5\times5$ kernels with padding $2$, and column-wise max pooling of the
		rotation-equivariant feature map to a $1\times16$ map that a dense
		layer maps to the $128$-D descriptor. Kernel size, padding
		(\textsf{p}) and non-default stride (\textsf{s}) are shown above each
		layer; channel count below.}
	\label{fig:network_architecture}
\end{figure}

\noindent\textbf{Describe.}~ \label{sec:describe}
Each canonical patch is embedded into a $128$-dimensional unit-norm descriptor
using a HardNet-style convolutional network~\cite{mishchuk2017working},
following the log-polar descriptor setting of
Ebel~\emph{et al.}~\cite{ebel2019beyond}. The network takes the $64\times64$
patch as input, uses $5\times5$ convolutions with padding $2$, and maps the
final feature map to a 128-D descriptor with a dense layer. Because we do not
assign a dominant orientation to the blobs, we max-pool along the angular axis
after the final convolution. The log-polar transform converts residual rotation
and isotropic scaling into translations along the angular and radial axes
respectively; angular max pooling makes the descriptor rotation-invariant.
Convolutions wrap circularly on the angular axis, which is periodic, and pad
at the radial edges, which are not; downsampling is antialiased. \Cref{fig:network_architecture} gives the full
architecture.

\noindent\textbf{Curriculum training.}~Training minimizes the supervised contrastive loss~\cite{khosla2020supervised}. Every blob on every board in the dataset is uniquely identified by its board ID and blob index, and each $(\text{board ID},,\text{blob index})$ pair defines a class. At each training stage, we optimize with AdamW~\cite{loshchilov2019decoupledweightdecayregularization} using a learning rate of $10^{-4}$ and a batch size of $4{,}096$, and select models by false-positive rate at $95\%$ recall (FPR95) on the validation set.

We first train for $1000$ epochs on synthetic images of $40$ dynamically generated boards composited over OpenLORIS~\cite{shi2019openlorisscene} backgrounds under random homographies~\cite{detone2017superpoint}, which provide exact board-to-image blob correspondences. Training images are augmented with Gaussian blur, Gaussian noise, and color jitter.

This preliminary network is then used within the RANSAC pipeline to bootstrap labels from $634$ indoor and outdoor videos and photographs of $50$ boards. For each image, the estimated board homography is used to retain spatially verified blob detections and assign each the class label of its corresponding reference blob; unmatched detections are retained as confusers. The boards are split $45/5$ into training and validation sets, approximately $90{:}10$ by patch count. No image from the motion-capture evaluation dataset is used at any stage of training, validation, or bootstrapping. A second network is trained for $50$ epochs on these data and used for one further bootstrap iteration. Random downsampling and homographic augmentation provide additional hard samples. The final network used for evaluation is trained on the resulting third-generation real-image patches.

Reference-pattern patches and their corresponding real observations share the same class label, so the reference-to-observation matches required at test time are explicit positive pairs during training. The batch-sampling procedure is detailed in \suppref{sec:supp:training}.

\subsection{Match, Verify, Certify \& Claim}
\label{sec:match_verify}
\noindent\textbf{Match.}~ For each board, canonical patches are extracted from the reference pattern using the
known blob geometry and their descriptors precomputed. This gallery is built
once and reused across queries, so it incurs no per-query descriptor cost.
Matching remains linear in gallery size: for $N$ image detections and $B$
boards with $M$ blobs each, one $N\times M$ cosine-similarity matrix is
computed per board, for $O(NBMD)$ cost with $D$-dimensional descriptors.
Verification fits at most one hypothesis per live board per round, giving
$O(BK)$ fits for the $K$ boards ultimately committed. Neither term is
quadratic in $B$.

Detections are matched against each board's gallery independently, by mutual
nearest neighbours in cosine similarity above a threshold, which suppresses
chance matches between locally similar neighbourhoods across boards. Matching
proceeds in two phases: a strict mutual-nearest-neighbour pass proposes
boards, and a permissive $k$-nearest-neighbour pass ($k{=}3$) later rematches
the detections that remain once boards have been accepted. Each candidate
board yields a set of
tentative detection-to-blob correspondences with associated descriptor
similarities.

\noindent\textbf{Spatial verification.}~ Each candidate correspondence set is
verified by RANSAC~\cite{fischler1981random}. Verification is agnostic to the geometric
model: any solver registering reference blobs to image detections can be used.
The principal cases are P3P~\cite{kneip2011p3p,persson2018lambda} for a
calibrated camera and a planar homography~\cite{Hartley00} when the camera is
uncalibrated. Each accepted model is refined on its inlier set using the
local-optimization step of LO-RANSAC~\cite{chum2003locally}, and hypotheses
with fewer than $N_{\mathrm{min}}$ final inliers are discarded.

\noindent\textbf{Appearance certification.}~ At BlobBoard densities, geometric
consensus alone is insufficient for reliable identity verification.
High-frequency image content, such as a tree, can generate many detections,
some of which may by chance be geometrically consistent with a proposed
registration to a gallery board. Requiring stronger geometric consensus would
suppress these false hypotheses but also reject difficult boards with few
usable correspondences. We therefore use a permissive geometric threshold and
certify identity from appearance: at least half of the geometric inliers must
have cosine similarity $\geq 0.6$ to their matched reference blobs. Each
inlier already names a reference blob and board, so this requires no
additional feature extraction or decoding. False hypotheses have median
per-inlier similarity $0.07$--$0.37$, compared with $0.73$--$0.97$ for genuine
boards.

\noindent\textbf{Claim.}~ Because multiple boards may compete for the same
detections, correspondences are claimed in rounds rather than all at once. In
each round, all live candidates
are fit, the best-supported board is committed, and detections within its
quadrilateral are claimed. The remaining candidates are then rematched
one-to-one against the unclaimed detections before the next round. Claiming
removes correspondences explained by an accepted board, preventing false
hypotheses from borrowing its support, while rematching recovers neighbouring
boards from the remaining features. The output is the set of verified
$(\text{board id},\,\text{pose})$ pairs passing the geometry, appearance, and
orientation tests.

\subsection{Planar pose ambiguity}
\label{sec:orientation}
A planar target admits a two-fold pose
ambiguity~\cite{schweighofer2006robust}: two poses with plane normals
reflected about the viewing direction can produce nearly identical
reprojections. Perspective separates the two branches when the board is large
in the image and strongly foreshortened, but the distinction becomes weak
near fronto-parallel views or at small projected extent. Appearance cannot
resolve this ambiguity either, because the rotation-invariant canonical
descriptor can certify essentially the same correspondences under both poses.
We therefore construct the competing mirror pose, refine both branches from
the same correspondences using identical local optimization, and retain the
branch with the higher scale-marginalized inlier/outlier
score~\cite{pritts2026scoring}. The score models inlier reprojection error as
Gaussian, marginalizes the unknown noise scale $\sigma^2$ in closed form
under a Jeffreys prior, and models outliers uniformly. For pose $\vtheta$
with inlier set $I$ among $N$ correspondences,
\begin{equation}
	S(\vtheta,I)=\log\Gamma\!\left(\tfrac{\nu}{2}\right)
	-\tfrac{\nu}{2}\log\!\Bigl(\pi\!\!\sum_{i\in I}\!r_i^2\Bigr)
	-d_g\,(N-|I|)\log 2a,
	\quad \nu=d_g|I|-n_{\vtheta},
	\label{eq:marginal-score}
\end{equation}
where $r_i$ is the reprojection residual, $d_g=2$ is the residual dimension
per correspondence, $n_{\vtheta}=6$ is the number of pose degrees of freedom,
and $a=5$ pixels is the outlier half-width. Because $\sigma^2$ is
marginalized, the score has no noise scale to tune. For each branch, the
maximizing inlier set $I$ is found exactly by sweeping the sorted
residuals~\cite{pritts2026scoring}. Since each branch is independently
optimized and scored with its own maximizing inlier set, their score
difference is a generalized log-likelihood ratio, of which only the sign is
used (\cref{tab:flips}). Thus geometry localizes the board, appearance
certifies its identity, and the likelihood ratio resolves its orientation.

\section{Evaluation Protocol}
\label{sec:protocol}

\begin{figure}[t]
	\centering
	\begin{minipage}{0.32\textwidth}
		\centering
		{\small BlobBoards}\\[2pt]
		\includegraphics[width=\textwidth]{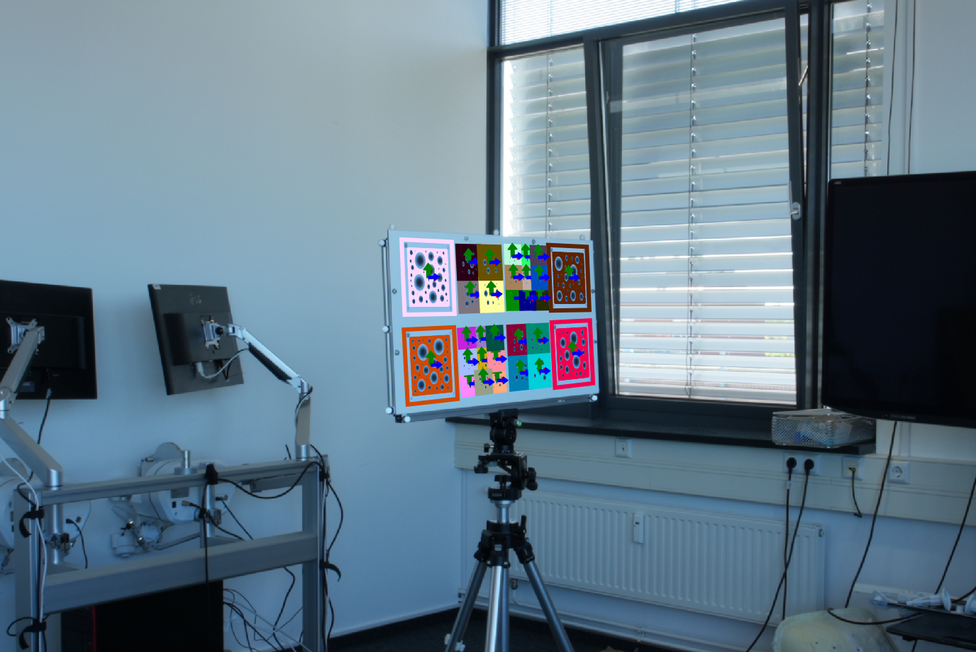}
	\end{minipage}\hfill
	\begin{minipage}{0.32\textwidth}
		\centering
		{\small AprilTag}\\[2pt]
		\includegraphics[width=\textwidth]{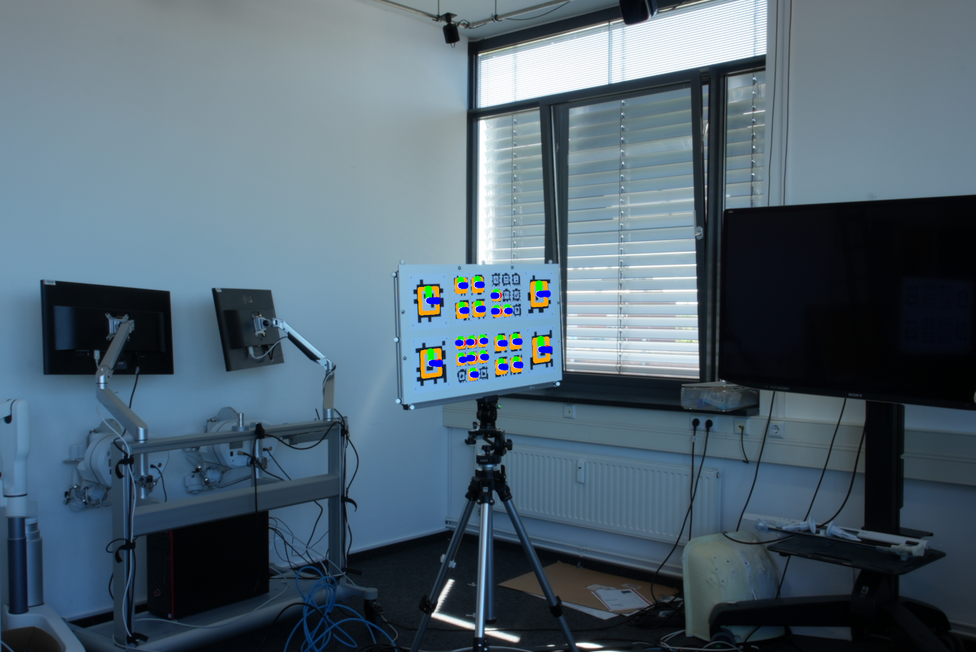}
	\end{minipage}\hfill
	\begin{minipage}{0.32\textwidth}
		\centering
		{\small ArUco}\\[2pt]
		\includegraphics[width=\textwidth]{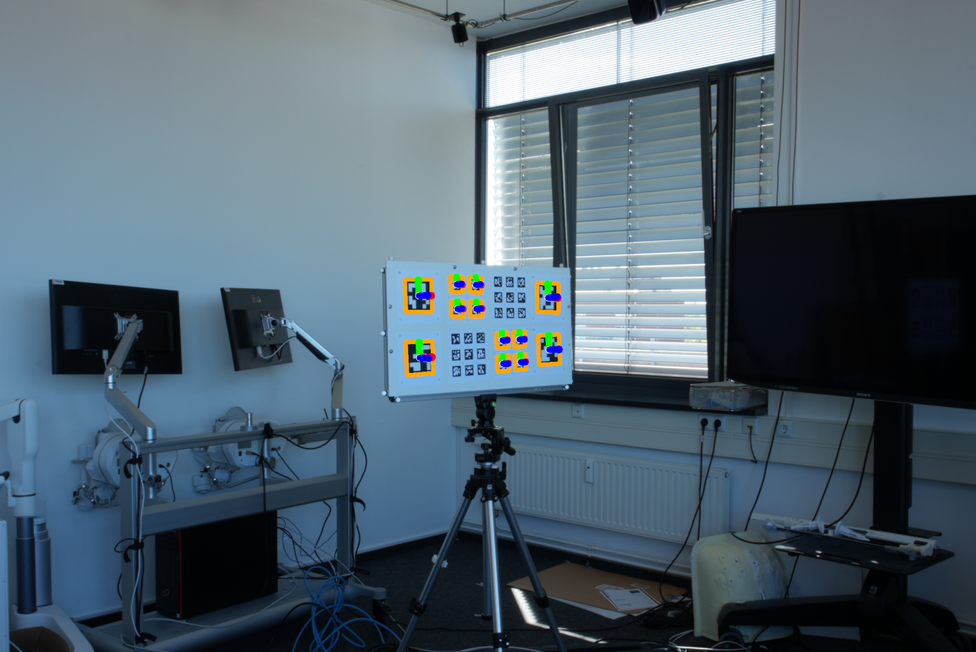}
	\end{minipage}
	\caption{Experimental setup, one capture per marker type. Each type is mounted
		on a matched rigid panel carrying $30$ markers at three physical scales
		($18$ small, $8$ medium, $4$ large), observed while motion capture
		tracks the panel's $6$-DoF pose. Estimated board frames are overlaid;
		for BlobBoards the reference pattern is additionally projected under
		the recovered pose. Further captures are in \suppref{fig:supp:setup}.}
	\label{fig:evaluation_setup_results}
\end{figure}


We compare BlobBoards with AprilTag~3
(\textit{tagStandard41h12})~\cite{olson2011apriltag,krogius2019flexible} and
OpenCV's ArUco
detector~\cite{garrido2014automatic,bradski2000opencv} with the ArUco~3
\textit{MIP\_36h12} dictionary~\cite{romero2018speeded}, the
state-of-the-art baselines for independently identifiable single markers
and the class to which this evaluation is scoped. Both recover pose from
their four detected corners on undistorted images with the calibrated
intrinsics, using OpenCV's \texttt{IPPE\_SQUARE}
solver~\cite{collins2014ippe}, which selects between the two planar poses
(\cref{sec:orientation}). All patterns are rendered at $1200$\,dpi on the same printer and
rigidly mounted on flat aluminium panels, each carrying $30$ markers at three
physical scales: $18$ small ($4$\,cm), $8$ medium ($6$\,cm) and $4$ large
($12$\,cm). Pattern area is matched across marker types at each scale, so every
method receives the same image area; what differs is the measurement density
inside it, a median of $213$, $375$ and $1027$ blobs at the three sizes
against four quad corners for a tag of any size (generation settings in
\suppref{tab:supp:params}).

A fixed DSLR observes the panel while an infrared
motion-capture system tracks rigid marker rigs on both the panel and the
camera. Captures are made at fixed camera distances of $1$, $2$, and $3$\,m
while rotating the panel in nominal $10^\circ$ increments from $-90^\circ$ to
$+90^\circ$ about fronto-parallel; the three marker types are captured at
matched distances and viewpoints (see~fig.~\ref{fig:evaluation_setup_results}). The actual obliquity of every frame is
obtained from the calibrated motion-capture pose (\cref{sec:ground_truth}).

\begin{figure}[!t]
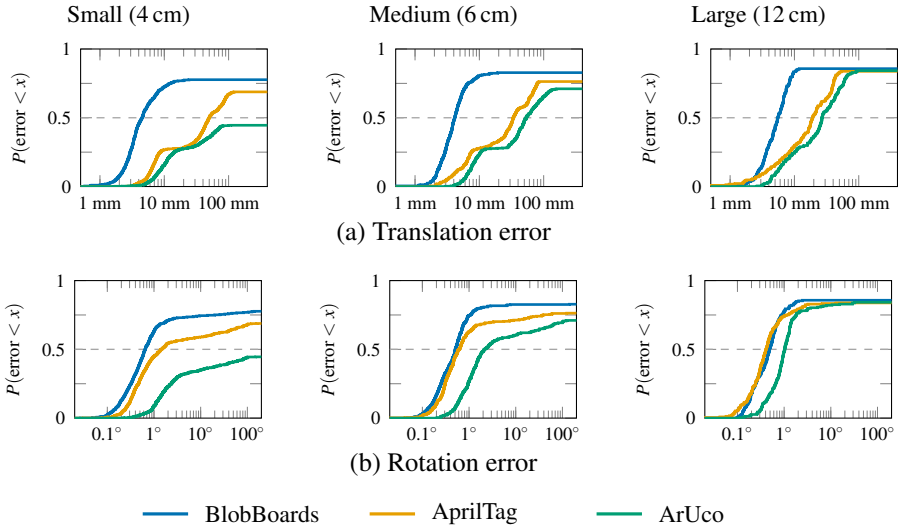

	\centering
\centering
\setlength{\tabcolsep}{2pt}
\providecolor{bbGreen}{HTML}{009E73}
\providecolor{bbBlue}{HTML}{0072B2}
\providecolor{bbOrange}{HTML}{E69F00}
\begin{tabular}{@{}ccc@{}}
{\small Small (4\,cm)} & {\small Medium (6\,cm)} & {\small Large (12\,cm)} \\[1pt]
\input{figures/floorplan/hist_trans_small} & \input{figures/floorplan/hist_trans_medium} & \input{figures/floorplan/hist_trans_large} \\
\multicolumn{3}{c}{(a) Translation error} \\[6pt]
\input{figures/floorplan/hist_rot_small} & \input{figures/floorplan/hist_rot_medium} & \input{figures/floorplan/hist_rot_large} \\
\multicolumn{3}{c}{(b) Rotation error} \\
\end{tabular}\\[4pt]
\begin{tikzpicture}[baseline]
\draw[bbBlue, line width=1.2pt] (0.0cm,0) -- (0.6cm,0);
\node[anchor=west, font=\small] at (0.7cm,0) {BlobBoards};
\draw[bbOrange, line width=1.2pt] (3.0cm,0) -- (3.6cm,0);
\node[anchor=west, font=\small] at (3.7cm,0) {AprilTag};
\draw[bbGreen, line width=1.2pt] (6.0cm,0) -- (6.6cm,0);
\node[anchor=west, font=\small] at (6.7cm,0) {ArUco};
\end{tikzpicture}

	\caption{Recall as a function of the error threshold, by marker size.
		Columns show small ($4$\,cm), medium ($6$\,cm), and large ($12$\,cm)
		markers; rows show translation and rotation error. A missed detection
		never enters the numerator, so each curve saturates at the
		corresponding probability of detection. Dashed lines mark recall
		$0.5$.}
	\label{fig:eval:scale_ecdf}
\end{figure}

\subsection{Ground-Truth Calibration}
\label{sec:ground_truth}
Motion capture provides the panel-rig-to-camera-rig pose
$\mathbf{A}_i$ for each image $i$. Converting this to the camera frame
requires two fixed transforms: the camera-rig-to-optical-frame
transform $\mathbf{X}$, shared by all $30$ boards, and the
board-to-panel-rig transform $\mathbf{Y}_m$, one per board. We estimate
these from the board pose $\hat{\mathbf{B}}_{im}$ returned by the marker
system:
\[
	\hat{\mathbf{B}}_{im}=\mathbf{X}\,\mathbf{A}_i\,\mathbf{Y}_m,
\]
an instance of robot--world/hand--eye calibration~\cite{shah2013solving}.
Relative motion between two frames of the same board cancels
$\mathbf{Y}_m$, allowing the rotation of $\mathbf{X}$ to be estimated
jointly from all boards. Each board rotation then follows from its own
frames, after which the translations of $\mathbf{X}$ and all
$\mathbf{Y}_m$ are recovered jointly by linear least squares
(\suppref{sec:supp:derivations}). The calibrated pose of each board is then
available for every image.

\noindent\textbf{Entangled errors.}~ The panel and camera mounts
introduce mechanical error, while the board poses
$\hat{\mathbf{B}}_{im}$ used for calibration are themselves estimated
by the marker system. Mechanical and detector errors are therefore
entangled in the recovered transforms. To limit this coupling,
calibration is performed once per marker system on a set of images disjoint from
evaluation, spanning poses easy to recover for both motion tracking system and marker detectors.
The fixed mount transforms can therefore absorb only
systematic, pose-independent bias, not pose-dependent detector error
(\suppref{sec:supp:handeye}). The calibration residual provides an
empirical estimate of the ground-truth uncertainty, and
\suppref{tab:handeye} reports it for all three marker systems.
Translation is well clear of it ($0.68$\,mm against detected-board
medians of $\bbTransMedSmall$--$\bbTransMedLarge$\,mm), whereas the
median rotation error of $\bbRotMed^\circ$ is within a factor of
two. Median rotation differences between the systems are therefore
smaller than this ground truth can resolve, and we compare rotation
only on the large flip errors of \cref{tab:flips}, which are one to
two orders of magnitude above the residual.

\begin{figure*}[t]
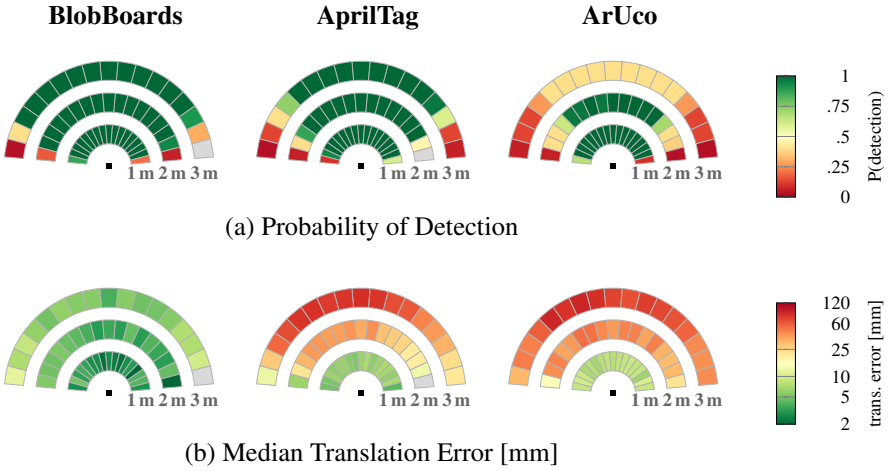

\providecommand{\fpscale}{1}\renewcommand{\fpscale}{0.42}   
\setlength{\tabcolsep}{3pt}
\centering
\begin{tabular}{@{}ccc l@{}}
\textbf{BlobBoards} & \textbf{AprilTag} & \textbf{ArUco} & \\[2pt]
\input{figures/floorplan/floorplan_blob_det} & \input{figures/floorplan/floorplan_april_det} & \input{figures/floorplan/floorplan_aruco_det} & 
\begin{tikzpicture}[baseline=(current bounding box.north)]
\pgfplotsset{colormap={rdylgnrev}{rgb=(0.0,0.408,0.216) rgb=(0.0729,0.5423,0.286) rgb=(0.2297,0.6581,0.3457) rgb=(0.4359,0.7567,0.392) rgb=(0.6151,0.8353,0.412) rgb=(0.7653,0.9001,0.4897) rgb=(0.8936,0.955,0.6033) rgb=(1.0,1.0,0.749) rgb=(0.9971,0.9129,0.6033) rgb=(0.9943,0.794,0.4743) rgb=(0.987,0.6456,0.3633) rgb=(0.962,0.4634,0.2797) rgb=(0.8919,0.2904,0.2001) rgb=(0.787,0.1343,0.1519) rgb=(0.647,0.0,0.149)}, colormap={rdylgn}{rgb=(0.647,0.0,0.149) rgb=(0.787,0.1343,0.1519) rgb=(0.8919,0.2904,0.2001) rgb=(0.962,0.4634,0.2797) rgb=(0.987,0.6456,0.3633) rgb=(0.9943,0.794,0.4743) rgb=(0.9971,0.9129,0.6033) rgb=(1.0,1.0,0.749) rgb=(0.8936,0.955,0.6033) rgb=(0.7653,0.9001,0.4897) rgb=(0.6151,0.8353,0.412) rgb=(0.4359,0.7567,0.392) rgb=(0.2297,0.6581,0.3457) rgb=(0.0729,0.5423,0.286) rgb=(0.0,0.408,0.216)}}
\begin{axis}[hide axis, scale only axis, width=0pt, height=1.6cm,
  colormap name=rdylgn, colorbar, point meta min=0.0, point meta max=1.0,
  colorbar style={height=1.6cm, width=0.25cm, ylabel={P(detection)}, ylabel style={font=\scriptsize},
    ytick={0.0,0.25,0.5,0.75,1.0}, yticklabels={0,.25,.5,.75,1}, tick label style={font=\scriptsize},
    yticklabel style={text width=1.7em, align=right}}]
  \addplot[draw=none] coordinates {(0,0)};
\end{axis}
\end{tikzpicture} \\[2pt]
\multicolumn{3}{c}{(a) Probability of Detection} & \\[8pt]
\input{figures/floorplan/floorplan_blob_trans} & \input{figures/floorplan/floorplan_april_trans} & \input{figures/floorplan/floorplan_aruco_trans} & 
\begin{tikzpicture}[baseline=(current bounding box.north)]
\pgfplotsset{colormap={rdylgnrev}{rgb=(0.0,0.408,0.216) rgb=(0.0729,0.5423,0.286) rgb=(0.2297,0.6581,0.3457) rgb=(0.4359,0.7567,0.392) rgb=(0.6151,0.8353,0.412) rgb=(0.7653,0.9001,0.4897) rgb=(0.8936,0.955,0.6033) rgb=(1.0,1.0,0.749) rgb=(0.9971,0.9129,0.6033) rgb=(0.9943,0.794,0.4743) rgb=(0.987,0.6456,0.3633) rgb=(0.962,0.4634,0.2797) rgb=(0.8919,0.2904,0.2001) rgb=(0.787,0.1343,0.1519) rgb=(0.647,0.0,0.149)}, colormap={rdylgn}{rgb=(0.647,0.0,0.149) rgb=(0.787,0.1343,0.1519) rgb=(0.8919,0.2904,0.2001) rgb=(0.962,0.4634,0.2797) rgb=(0.987,0.6456,0.3633) rgb=(0.9943,0.794,0.4743) rgb=(0.9971,0.9129,0.6033) rgb=(1.0,1.0,0.749) rgb=(0.8936,0.955,0.6033) rgb=(0.7653,0.9001,0.4897) rgb=(0.6151,0.8353,0.412) rgb=(0.4359,0.7567,0.392) rgb=(0.2297,0.6581,0.3457) rgb=(0.0729,0.5423,0.286) rgb=(0.0,0.408,0.216)}}
\begin{axis}[hide axis, scale only axis, width=0pt, height=1.6cm,
  colormap name=rdylgnrev, colorbar, point meta min=0.301, point meta max=2.0792,
  colorbar style={height=1.6cm, width=0.25cm, ylabel={trans.~error [mm]}, ylabel style={font=\scriptsize},
    ytick={0.301,0.699,1.0,1.3979,1.7782,2.0792}, yticklabels={2,5,10,25,60,120}, tick label style={font=\scriptsize},
    yticklabel style={text width=1.7em, align=right}}]
  \addplot[draw=none] coordinates {(0,0)};
\end{axis}
\end{tikzpicture} \\[2pt]
\multicolumn{3}{c}{(b) Median Translation Error [mm]} & \\
\end{tabular}

	\caption{Detection and pose accuracy over the capture floor plan.
		Concentric rings denote camera distance ($1$--$3$\,m) and wedges denote
		nominal $10^\circ$ viewing-angle stations. Colour encodes (a)
		probability of detection and (b) median translation error. Grey wedges
		indicate visited stations with no detections.}
	\label{fig:floorplan}
\end{figure*}

\begin{figure}[t]
	\centering
	\begin{minipage}[c]{0.19\linewidth}
		\includegraphics[width=\linewidth]{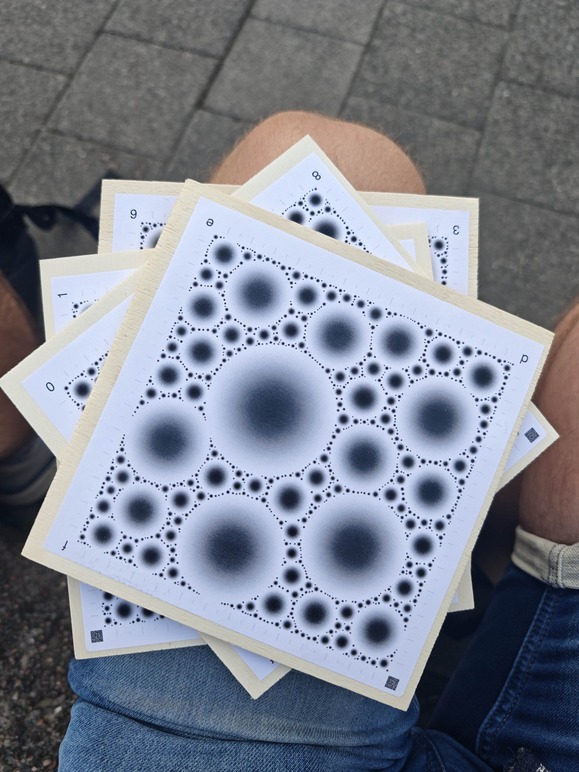}
	\end{minipage}\hfill
	\begin{minipage}[c]{0.19\linewidth}
		\includegraphics[width=\linewidth]{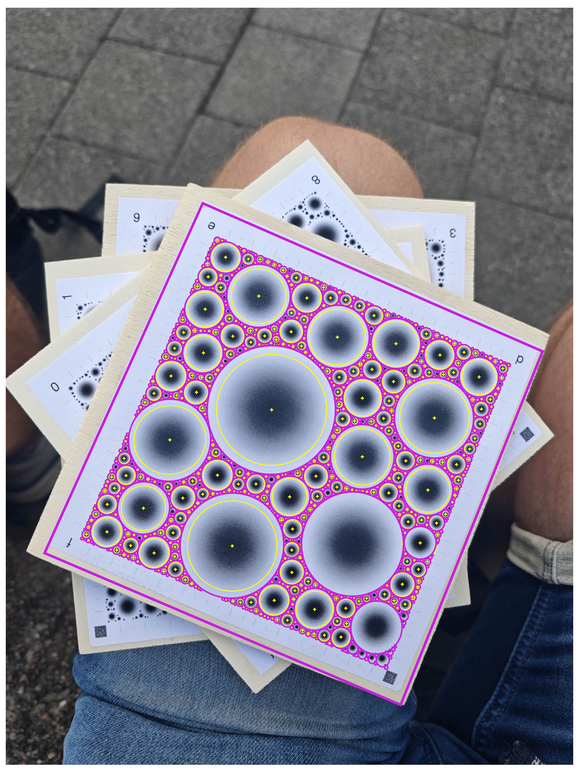}
	\end{minipage}\hfill
	\begin{minipage}[c]{0.19\linewidth}
		\includegraphics[width=\linewidth]{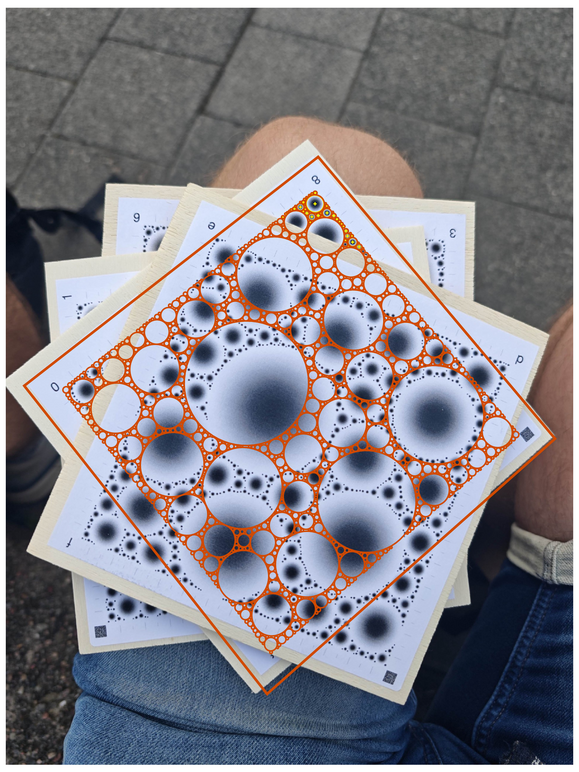}
	\end{minipage}\hfill
	\begin{minipage}[c]{0.19\linewidth}
		\includegraphics[width=\linewidth]{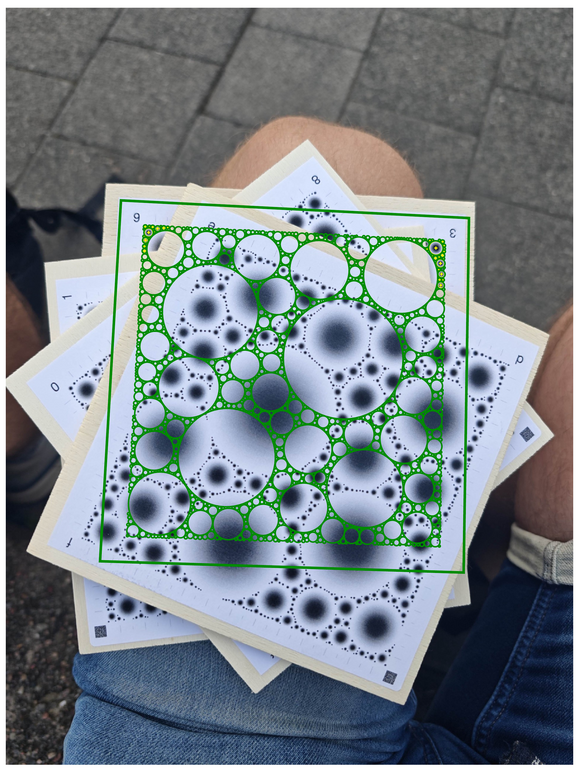}
	\end{minipage}\hfill
	\begin{minipage}[c]{0.19\linewidth}
		\includegraphics[width=\linewidth]{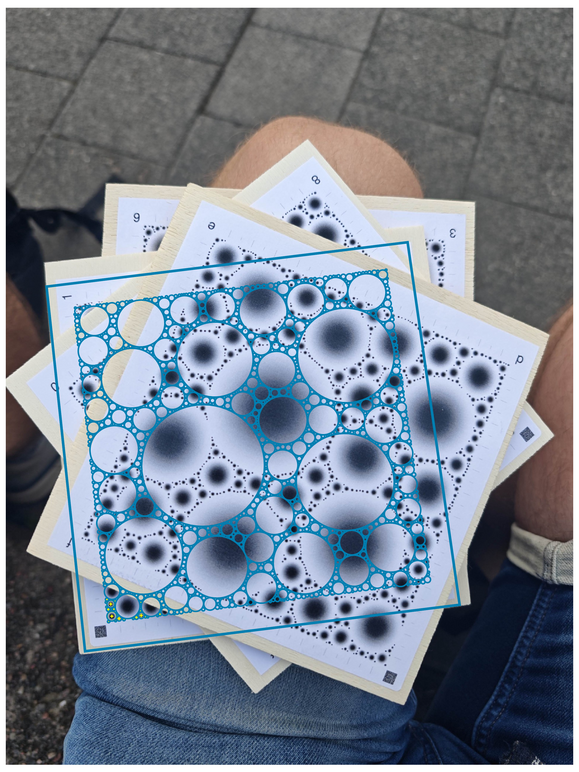}
	\end{minipage}\\[3pt]
	\begin{minipage}[c]{0.31\linewidth}
		\includegraphics[width=\linewidth]{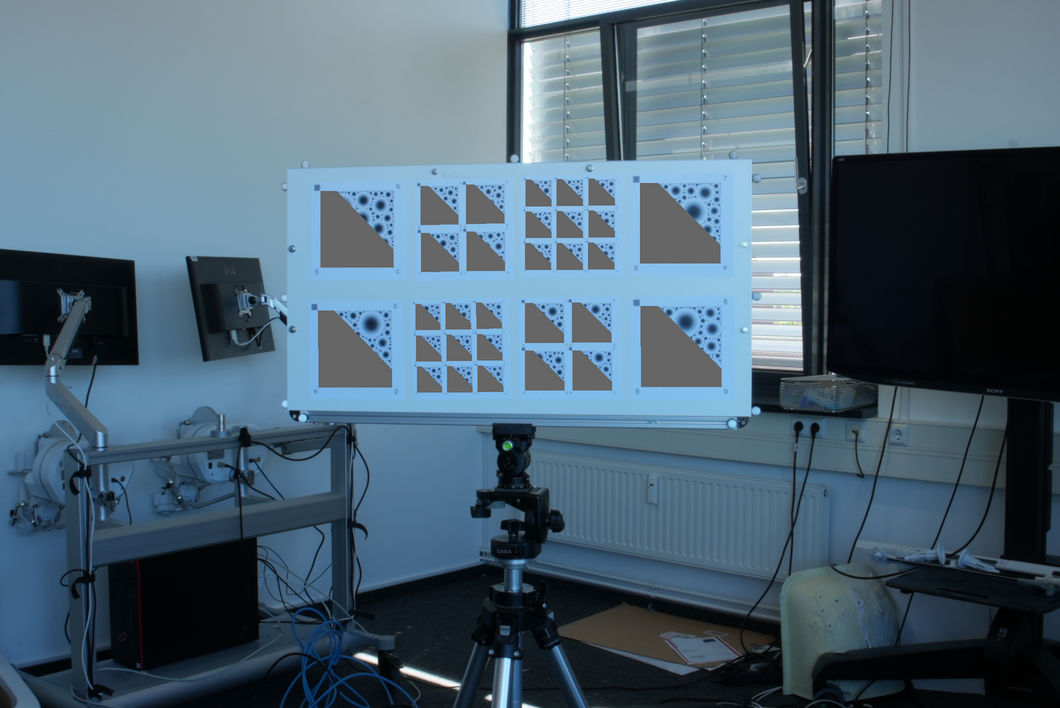}
	\end{minipage}\hfill
	\begin{minipage}[c]{0.31\linewidth}
		\includegraphics[width=\linewidth]{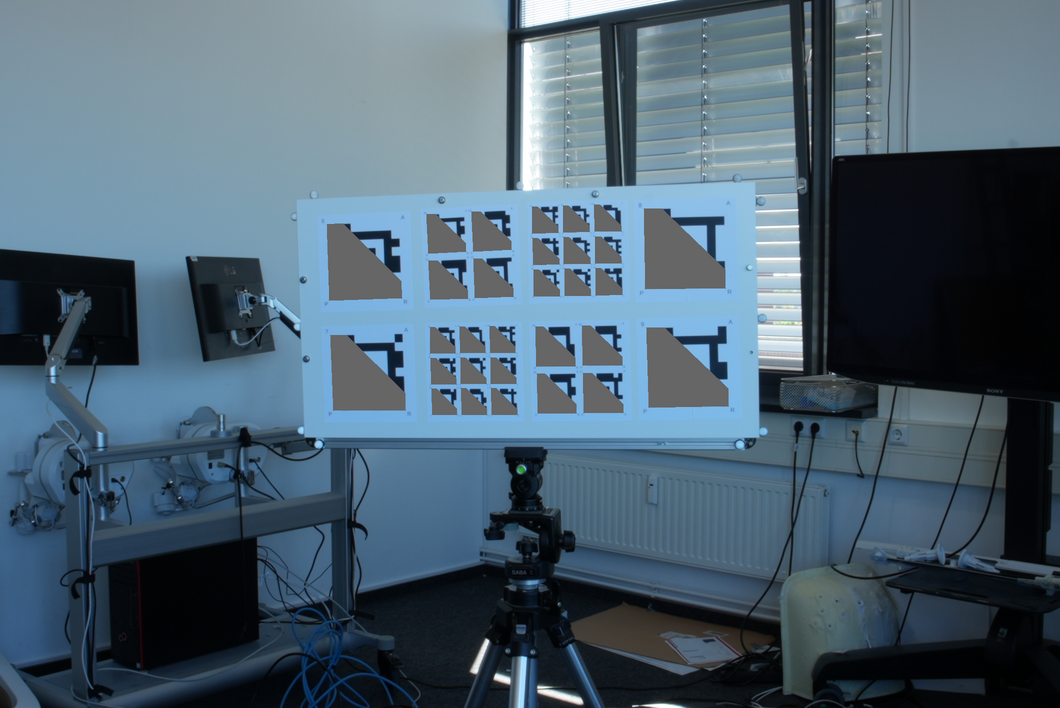}
	\end{minipage}\hfill
	\begin{minipage}[c]{0.31\linewidth}
		\includegraphics[width=\linewidth]{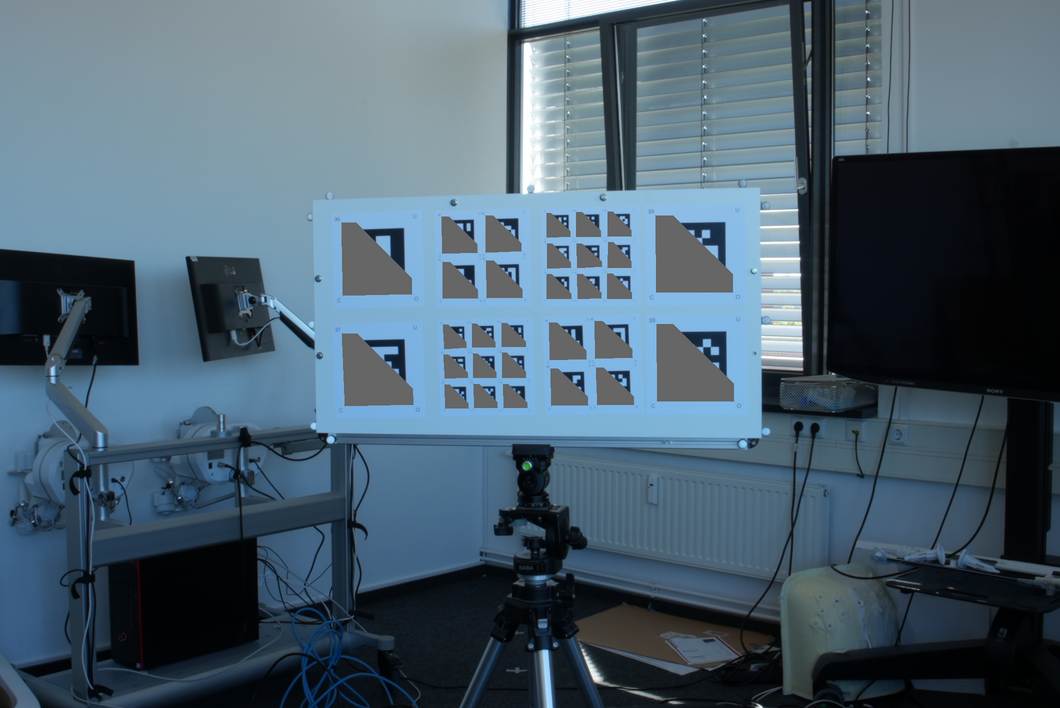}
	\end{minipage}
	\caption{Occlusion examples. \emph{Top:} detected BlobBoards under
		physical occlusion, shown with recovered board projections and
		affine-adapted feature ellipses. \emph{Bottom:} the synthetic $70\%$
		occlusion mask applied to BlobBoards, AprilTag, and ArUco. Masks are
		defined in board coordinates and warped using the ground-truth
		homography, giving matched occlusion across methods and viewpoints.}
	\label{fig:occexamples}
\end{figure}

\section{Results}
\label{sec:results}
We evaluate probability of detection, absolute pose accuracy, and
robustness to occlusion. Translation and rotation are reported
separately because they respond differently to marker size and
viewpoint. We report two complementary quantities. \emph{Recall} at
error threshold $x$, written $P(\mathrm{error}<x)$, is the fraction of
all visible board-viewing attempts that are detected, correctly
identified, and posed with error below $x$. Missed detections therefore
count against the method at every finite threshold, preventing a
detector from appearing accurate by declining difficult cases.
\emph{Summary statistics}, including medians and percentiles, are
computed over correctly detected boards only and isolate pose accuracy
after successful recovery.

\Cref{fig:eval:scale_ecdf} plots $P(\mathrm{error}<x)$ for translation
and rotation at each marker size, equivalently the empirical CDF of
error over all attempts with failures assigned infinite error;
\suppref{fig:supp:pod_trans} pools the three sizes. Each
curve saturates at the method's \emph{probability of detection} (PoD),
the fraction of attempts that are detected and correctly identified,
and hence its maximum recall
$\lim_{x\to\infty} P(\mathrm{error}<x)$.

For every correctly identified board, we compare the estimated
camera-frame pose with motion-capture ground truth. Translation error is
the Euclidean distance between estimated and ground-truth board origins;
rotation error is the geodesic angle of the residual rotation. Because
marker dimensions and the motion-capture reconstruction are metric, translation
error is reported directly in millimetres.

\noindent\textbf{Translation.}~ BlobBoards maintain detected-board median translation errors of
$\bbTransMedSmall$, $\bbTransMedMedium$, and $\bbTransMedLarge$\,mm for
the small, medium, and large boards respectively. 
In contrast, the tag baselines
degrade strongly as the marker shrinks and with increasing viewing
distance and obliquity. Relative to AprilTag, BlobBoards reduce
translation error by $\gainAprSmall\times$ on small boards,
$\gainAprMedium\times$ on medium boards, and $\gainAprLarge\times$ on
large boards. Over all attempts, misses included, the $50\%$-recall
translation errors are $\bbAttTransMed$, $\aprAttTransMed$, and
$\ucoAttTransMed$\,mm for BlobBoards, AprilTag, and ArUco. The gain grows
as the marker shrinks, as the geometry predicts: a tag supplies four corner
measurements whatever its size, whereas a BlobBoard's constraints scale with
its blob count and remain well distributed as the projection is
foreshortened.

\noindent\textbf{Rotation.}~ BlobBoards also produce the tightest rotation-error distribution. 
Their median detected-board rotation error is $\bbRotMed^\circ$, compared with
$\aprRotMed^\circ$ for AprilTag and $\ucoRotMed^\circ$ for ArUco, and
the $50\%$-recall rotation errors over all attempts are
$\bbAttRotMed^\circ$, $\aprAttRotMed^\circ$, and $\ucoAttRotMed^\circ$. The difference from AprilTag lies primarily in the
tails: AprilTag's $90$th-percentile rotation error grows from
$\aprRotPninetyLarge^\circ$ on large boards to $\aprRotPninetySmall^\circ$ on
small ones, whereas the BlobBoard $90$th percentile stays between
$\bbRotPninetyLarge^\circ$ and $\bbRotPninetySmall^\circ$
; the typical errors of the two methods are
similar; those rotation tails are the planar pose ambiguity of \cref{sec:orientation}.

\noindent\textbf{Detection.}~ Because the panel geometry, board
identities, and motion-capture pose are known, every visible marker is
an attempt.  The evaluation gallery contains the panel's own $30$
boards, all present in every capture, so identification is a
$1$-of-$30$ assignment without absent distractors. Across $56$
captures, BlobBoards returned $\bbDetPoses$ poses, each with a unique gallery
identity within its capture. Misassignments are directly detectable
from the rigid panel layout: any swap would displace a board by at
least the $\bbMinSpacingMM$\,mm minimum centre spacing, an order of
magnitude beyond BlobBoards' median absolute translation error of
$\bbAttTransMed$\,mm, so no consistent pose can survive a swap. Thus, no
reported detection is a misidentification, and the appearance
certification of \cref{sec:match_verify} admitted no false board.

BlobBoards achieve the highest overall PoD, $\bbDet\%$, compared with
$\aprDet\%$ for AprilTag and $\ucoDet\%$ for ArUco. The advantage is
largest for the smallest markers, with detection rates of
$\bbDetSmall\%$, $\aprDetSmall\%$, and $\ucoDetSmall\%$,
respectively. All methods degrade toward grazing views, where
BlobBoards and AprilTag have comparable coverage at the most extreme
angles; \cref{fig:floorplan} shows the station-wise variation, and
\suppref{fig:supp:unoccluded} the same measurements as marginals against
obliquity and distance, with interquartile bands.

For BlobBoards, most remaining failures under severe foreshortening
occur after feature detection: blobs are still detected and
affine-adapted, but descriptor matching becomes less reliable where
real training tracks are sparse. Pose accuracy remains strong for
successfully identified boards.

Identification is challenging because each image contains up to $30$
markers of the same type, matched jointly against a gallery of
visually similar random patterns. Spatial verification resolves local
descriptor ambiguities by requiring a common pose, while appearance
certification rejects geometrically plausible hypotheses formed from
unrelated detections.

\begin{figure}[t]
	\centering
\centering
\setlength{\tabcolsep}{2pt}
\providecolor{bbGreen}{HTML}{009E73}
\providecolor{bbBlue}{HTML}{0072B2}
\providecolor{bbOrange}{HTML}{E69F00}
\begin{tabular}{@{}ccc@{}}
\providecolor{bbGreen}{HTML}{009E73}
\providecolor{bbBlue}{HTML}{0072B2}
\providecolor{bbOrange}{HTML}{E69F00}
\begin{tikzpicture}[baseline=(current bounding box.center)]
\begin{axis}[width=4.05cm, height=3.4cm, xmin=-5, xmax=95,
  ylabel={median error ($^\circ$)}, xlabel={occlusion level},
  xtick={0,30,60,90}, xticklabels={{0\%},{30\%},{60\%},{90\%}},
  ymode=log, log ticks with fixed point, ytick={0.5,1,2,5,10,25},
  label style={font=\scriptsize}, tick label style={font=\scriptsize},
  ]
\addplot[draw=bbBlue, line width=1.0pt, mark=*, mark size=1.1pt, mark options={fill=bbBlue, draw=bbBlue}] coordinates {(0,1.077) (10,1.103) (20,1.112) (30,1.147) (40,1.061) (50,1.054) (60,0.8871) (70,0.89) (80,0.7415) (90,0.7878)};
\addplot[draw=bbOrange, line width=1.0pt, mark=*, mark size=1.1pt, mark options={fill=bbOrange, draw=bbOrange}] coordinates {(0,1.242) (10,1.617)};
\addplot[draw=bbGreen, line width=1.0pt, mark=*, mark size=1.1pt, mark options={fill=bbGreen, draw=bbGreen}] coordinates {(0,3.036) (10,18.79) (20,24.24)};
\end{axis}
\end{tikzpicture} & 
\providecolor{bbGreen}{HTML}{009E73}
\providecolor{bbBlue}{HTML}{0072B2}
\providecolor{bbOrange}{HTML}{E69F00}
\begin{tikzpicture}[baseline=(current bounding box.center)]
\begin{axis}[width=4.05cm, height=3.4cm, xmin=-5, xmax=95,
  ylabel={median error (mm)}, xlabel={occlusion level},
  xtick={0,30,60,90}, xticklabels={{0\%},{30\%},{60\%},{90\%}},
  ymode=log, log ticks with fixed point, ytick={5,10,25,50,100},
  label style={font=\scriptsize}, tick label style={font=\scriptsize},
  ]
\addplot[draw=bbBlue, line width=1.0pt, mark=*, mark size=1.1pt, mark options={fill=bbBlue, draw=bbBlue}] coordinates {(0,3.757) (10,3.854) (20,3.872) (30,3.905) (40,3.991) (50,3.946) (60,3.776) (70,3.591) (80,3.569) (90,3.661)};
\addplot[draw=bbOrange, line width=1.0pt, mark=*, mark size=1.1pt, mark options={fill=bbOrange, draw=bbOrange}] coordinates {(0,25.62) (10,33.0)};
\addplot[draw=bbGreen, line width=1.0pt, mark=*, mark size=1.1pt, mark options={fill=bbGreen, draw=bbGreen}] coordinates {(0,19.04) (10,91.21) (20,107.6)};
\end{axis}
\end{tikzpicture} & 
\providecolor{bbGreen}{HTML}{009E73}
\providecolor{bbBlue}{HTML}{0072B2}
\providecolor{bbOrange}{HTML}{E69F00}
\begin{tikzpicture}[baseline=(current bounding box.center)]
\begin{axis}[width=4.05cm, height=3.4cm, xmin=-5, xmax=95,
  ylabel={$P(\mathrm{detection})$}, xlabel={occlusion level},
  xtick={0,30,60,90}, xticklabels={{0\%},{30\%},{60\%},{90\%}},
  ymin=0, ymax=1.05, ytick={0,0.25,0.5,0.75,1}, yticklabels={0,,0.5,,1},
  label style={font=\scriptsize}, tick label style={font=\scriptsize},
  ]
\addplot[draw=bbBlue, line width=1.0pt, mark=*, mark size=1.1pt, mark options={fill=bbBlue, draw=bbBlue}] coordinates {(0,0.8012) (10,0.8018) (20,0.7994) (30,0.7887) (40,0.7542) (50,0.6911) (60,0.5923) (70,0.4851) (80,0.3595) (90,0.2226)};
\addplot[draw=bbOrange, line width=1.0pt, mark=*, mark size=1.1pt, mark options={fill=bbOrange, draw=bbOrange}] coordinates {(0,0.7281) (10,0.1099) (20,0.0005848) (30,0.0005848) (40,0.0005848) (50,0.0005848) (60,0.0005848) (70,0.0005848) (80,0.0005848) (90,0.0005848)};
\addplot[draw=bbGreen, line width=1.0pt, mark=*, mark size=1.1pt, mark options={fill=bbGreen, draw=bbGreen}] coordinates {(0,0.569) (10,0.2304) (20,0.1023) (30,0.002924) (40,0.004094) (50,0.0) (60,0.0) (70,0.0) (80,0.0) (90,0.0)};
\end{axis}
\end{tikzpicture} \\[1pt]
(a) Rotation error & (b) Translation error & (c) Probability of detection \\
\end{tabular}\\[4pt]
\begin{tikzpicture}[baseline]
\draw[bbBlue, line width=1.2pt] (0.0cm,0) -- (0.6cm,0);
\node[anchor=west, font=\small] at (0.7cm,0) {BlobBoards};
\draw[bbOrange, line width=1.2pt] (3.0cm,0) -- (3.6cm,0);
\node[anchor=west, font=\small] at (3.7cm,0) {AprilTag};
\draw[bbGreen, line width=1.2pt] (6.0cm,0) -- (6.6cm,0);
\node[anchor=west, font=\small] at (6.7cm,0) {ArUco};
\end{tikzpicture}

	\caption{Performance over the synthetic occlusion sweep for BlobBoards,
		AprilTag, and ArUco: \textbf{(a)}~median
		rotation error, \textbf{(b)}~median translation error, and
		\textbf{(c)}~probability of detection. Medians are omitted when fewer
		than $20$ detections survive at an occlusion level, which is why the
		tag curves stop early.}
	\label{fig:eval:occlusion}
\end{figure}
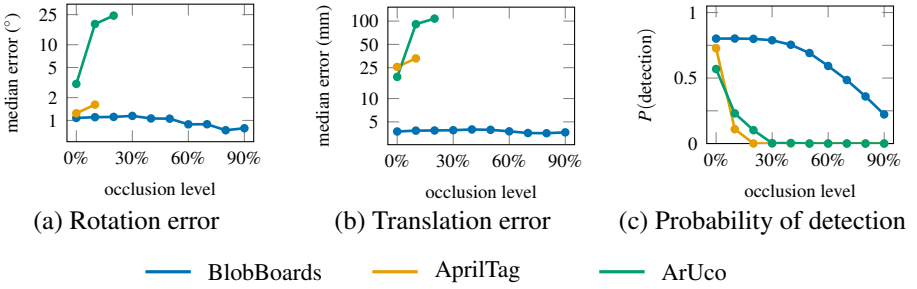

\noindent\textbf{Robustness to occlusion.}~ We evaluate occlusion by masking a
controlled fraction of each marker in reference coordinates and warping the
mask into the image with the ground-truth homography, ensuring the same
occluded area across marker types, scales, and viewpoints. Each mask is a
deterministic region cut by a single $45^\circ$ line anchored at a pattern
corner and offset to match the target area exactly (\cref{fig:occexamples};
construction in \suppref{sec:supp:occexamples}). Because the occluder always
crosses the marker border, the tag results characterize sensitivity to
border-crossing occlusion rather than an average over occluder placements.

The effect is immediate. At only $10\%$ occlusion, AprilTag PoD falls from
$\aprDet$ to $\aprDetOccTen$ and ArUco to $\ucoDetOccTen$, while BlobBoards
remain essentially unchanged at $\bbDetOccTen$. The surviving tag estimates
also degrade sharply: ArUco reaches $\ucoRotMedOccTen^\circ$ and
$\ucoTransMedOccTen$\,mm, and AprilTag $\aprTransMedOccTen$\,mm, compared with
$\bbRotMedOccTen^\circ$ and $\bbTransMedOccTen$\,mm for BlobBoards. From
$20\%$ occlusion onward AprilTag reports no detections, and ArUco reaches zero
by $30\%$.

BlobBoards instead degrade gradually. At $50\%$ occlusion they still
detect $\bbDetPctOccFifty\%$ of markers, with median errors of
$\bbRotMedOccFifty^\circ$ and $\bbTransMedOccFifty$\,mm;
\Suppref{fig:supp:occlusion50} shows that this coverage extends across
much of the capture floor plan, with \cref{fig:supp:occlusion10} and
\cref{fig:supp:occlusion70} giving the $10\%$ and $70\%$ levels of the
same series. Some poses are recovered even at $90\%$ occlusion.
Moreover, translation accuracy of successful detections remains nearly
constant across the sweep, with median error between
$\bbTransMedOccMin$ and $\bbTransMedOccMax$\,mm
(\cref{fig:eval:occlusion}). Occlusion removes correspondences but
poses remain accurate for BlobBoards.

\noindent\textbf{Planar pose ambiguity.}~ The rotation tails are the two-fold planar
pose ambiguity~\cite{schweighofer2006robust} of \cref{sec:orientation}, which
\cref{tab:flips} isolates at $>10^\circ$ rotation error. AprilTag produces
$\flipApr$ such errors and ArUco $\flipUco$, against $\flipBBraw$ for BlobBoards
before the mirror test and $\flipBBlrt$ after --- the likelihood-ratio test corrects
$\flipBBcorrected$ otherwise identical estimates without changing their detections,
correspondences or calibration. The tag baselines already use the
state-of-the-art planar solver IPPE~\cite{collins2014ippe}; BlobBoards decide the
same branch from hundreds of verified correspondences and can therefore still
improve on it.

\begin{table}[!t]
	\centering
\setlength{\tabcolsep}{10pt}
\begin{tabular}{@{}lcccc@{}}
\toprule
flips ($>10^\circ$) & \multicolumn{1}{c}{AprilTag} & \multicolumn{1}{c}{ArUco} & \multicolumn{1}{c}{BlobBoards} & \multicolumn{1}{c}{BlobBoards+LRT} \\
\midrule
$1\,\mathrm{m}$ & 3 & 21 & 0 & 0 \\
$2\,\mathrm{m}$ & 27 & 96 & 4 & 2 \\
$3\,\mathrm{m}$ & 94 & 35 & 41 & 31 \\
\midrule
total & 124 & 152 & 45 & 33 \\
\bottomrule
\end{tabular}

	\caption{Planar-pose flips (rotation error $>10^\circ$ against
		motion-capture ground truth) by tripod distance. IPPE selects the
		lower-reprojection-error tag pose from four corners, whereas
		\cref{sec:orientation} selects between BlobBoard branches using
		hundreds of verified correspondences. The final two columns use
		identical detections, correspondences, and calibration; the likelihood-ratio test (LRT)
		reduces the remaining flips from $\flipBBraw$ to $\flipBBlrt$.}
	\label{tab:flips}
\end{table}

\section{Implementation}
BlobBoards is implemented in Julia, with affine-warp detection, blob
adaptation and descriptor extraction as batched GPU kernels. We report
implementation timings rather than a hardware-normalized comparison, since
BlobBoards uses a GPU and both baselines run on the CPU. Measured warm on the
$12$-megapixel evaluation images containing $30$ boards, BlobBoards requires
$\bbTimeMeanS\pm\bbTimeStdS$\,s per image on an NVIDIA RTX 4070; AprilTag
requires $\aprTimeMeanS\pm\aprTimeStdS$\,s and ArUco
$\ucoTimeMeanS\pm\ucoTimeStdS$\,s on the CPU.
The cost is dominated by the dense multi-warp scale space, which is computed
once per image and shared across all candidate boards, so it is amortized over
the $30$ boards rather than paid per board.

The baselines are implemented in Python. AprilTag~3 runs through
\texttt{dt-apriltags} with \texttt{quad\_decimate} $2$ and edge refinement on;
ArUco runs through \texttt{cv.aruco} with default detector parameters. Both
are posed by \texttt{cv.solvePnP} with
\texttt{SOLVEPNP\_\allowbreak IPPE\_\allowbreak SQUARE}. All
RANSAC fits are seeded per board, so every figure and number here regenerates
bit-exactly from the published dataset record and the released configuration;
\suppref{tab:supp:settings} lists the detection, matching and verification
settings, and the evaluation scripts are released with the dataset.

\section{Conclusion}
\label{sec:conclusion}
BlobBoards cast marker detection and identification as feature matching,
spatial verification, and appearance certification rather than quad
segmentation, making measurement density a property of the printed pattern
rather than marker size. Gaussian blobs provide scale-covariant localization
with scale-independent peak response and admit a closed-form correction for
finite-support bias. Each blob is affine-adapted by a robust fit, while the
two-fold planar pose ambiguity is resolved by a scale-marginalized likelihood
test.

Against motion-capture ground truth, BlobBoards outperform state-of-the-art tag
systems in detection rate and translation accuracy while substantially reducing
the large-rotation failures caused by planar pose ambiguity. The gains are
largest for the smallest, most oblique, and most occluded boards, where
conventional tags are most fragile. Because a BlobBoard is constrained by
hundreds of features distributed across its printed area rather than four
corners, loss of visible area removes measurements progressively, allowing
identity and pose estimation to degrade gracefully rather than fail abruptly.\\

\makeatletter\ifbmv@review\else
	\noindent\textbf{Acknowledgements.}~ James Pritts is funded by
	Kiel Training for Excellence, an EU Horizon Europe programme under the
	Marie Sk\l{}odowska-Curie Actions (MSCA), grant agreement No~101081480. The motion-capture setup was funded through the project
	\emph{OP der Zukunft} as part of the Recovery Assistance for Cohesion and
	the Territories of Europe (REACT-EU) program and provided by the Kurt Semm
	Centre for laparoscopic and robot assisted surgery at the University
	Hospital of Schleswig-Holstein.
\fi\makeatother

\bibliography{egbib}

\clearpage
\setcounter{section}{0}
\setcounter{figure}{0}
\setcounter{table}{0}
\setcounter{equation}{0}
\setcounter{algorithm}{0}
\renewcommand{\thesection}{S\arabic{section}}
\renewcommand{\thefigure}{S\arabic{figure}}
\renewcommand{\thetable}{S\arabic{table}}
\renewcommand{\theequation}{S\arabic{equation}}
\renewcommand{\thealgorithm}{S\arabic{algorithm}}
\makeatletter
\let\arxiv@section\section
\renewcommand{\section}{\FloatBarrier\arxiv@section}
\makeatother
\renewcommand{\maketitle}{}
\section*{Supplementary Material}
\noindent The supplementary material of the paper follows in full. Its
sections, figures, tables, equations and algorithms carry the S prefix under
which the paper cites them.
\setlength{\emergencystretch}{3pt}
\maketitle

\section{Board Generation}
\label{sec:supp:generation}

\Cref{alg:gen} states the placement procedure. Blobs are placed largest-first
by rejection sampling with disjoint supports; the coarse-to-fine order matters,
because placing small blobs first leaves no room for large ones and yields a
substantially sparser board.

\begin{algorithm}[!ht]
	\caption{Greedy coarse-to-fine board generation: blobs are placed
		largest-first by rejection sampling with disjoint supports, then
		rasterised in a single pass.}
	\label{alg:gen}
	\begin{algorithmic}[1]
		\Require board size $W\!\times\!H$; scale schedule
		$\sigma_1>\sigma_2>\dots>\sigma_n$ with target counts $m_i$; seed
		\Ensure pattern image $P$; blob set $\mathcal{B}=\{(\mathbf{c}_j,\sigma_j)\}$
		\State $\mathcal{T}\gets$ empty R-tree of placed supports;\ \ $\mathcal{B}\gets\varnothing$
		\For{$i=1,\dots,n$ \textbf{(coarse-to-fine)} and $k=1,\dots,m_i$}
		\Comment{largest blobs first}
		\Repeat
		\State sample centre $\mathbf{c}$ uniformly within the inset board
		\If{support disc of $(\mathbf{c},\sigma_i)$ overlaps no blob in $\mathcal{T}$}
		\State insert $(\mathbf{c},\sigma_i)$ into $\mathcal{T}$ and $\mathcal{B}$
		\EndIf
		\Until{placed \textbf{or} trial budget exhausted}
		\Comment{skip if no free spot found}
		\EndFor
		\State $P\gets\mathbf{1}_{H\times W}$
		\Comment{white background: the $\|\mathbf{x}\|\!\to\!c\sigma$ limit of the blob profile}
		\State rasterise every blob in $\mathcal{B}$ over its support
		\Comment{disjoint supports $\Rightarrow$ single pass}
		\State \Return $P,\ \mathcal{B}$
	\end{algorithmic}
\end{algorithm}

\section{Board Generation Settings}
\label{sec:supp:params}

\Cref{tab:supp:params} lists the pattern parameters used to generate every
BlobBoard in the evaluation. All three physical sizes share one parameter set;
only the pattern extent differs, so the number of blobs a board carries grows
with its area. A tag of the same physical size contributes four corners at
every size.

\begin{table}[!h]
	\centering\small
	\begin{tabular}{@{}lccc@{}}
		\toprule
		                                   & small     & medium   & large    \\
		\midrule
		pattern extent (mm)                & $40$      & $60$     & $120$    \\
		blobs per board (median)           & $213$     & $375$    & $1027$   \\
		\midrule
		\multicolumn{4}{@{}l}{\emph{shared across all sizes}}                \\
		minimum blob scale $s_{\min}$      & \multicolumn{3}{c}{$0.2$\,mm}   \\
		maximum diameter fraction $d_f$    & \multicolumn{3}{c}{$0.5$}       \\
		support cutoff $c$          & \multicolumn{3}{c}{$2$}         \\
		scale-sampling distribution $\alpha$       & \multicolumn{3}{c}{$1.6$}       \\
		scales per octave                  & \multicolumn{3}{c}{$3$}         \\
		print density                      & \multicolumn{3}{c}{$1200$\,dpi} \\
		polarity                           & \multicolumn{3}{c}{dark}        \\
		\bottomrule
	\end{tabular}
	\caption{Pattern-generation settings for the evaluation boards ($18$ small,
		$8$ medium, $4$ large). Each board is generated from its own random
		seed, so blob counts vary slightly within a size class; the ranges are
		$202$--$226$, $365$--$390$, and $1001$--$1032$.}
	\label{tab:supp:params}
\end{table}

\section{Finite-Support Scale Correction}
\label{sec:supp:scalecorr}

Truncating the Gaussian at $c$ and removing the pedestal
$p=e^{-c^2/2}$ perturbs the scale-space response. Writing
$\rho=\sigma/\sigma_0$, $\kappa=c^2(1+\rho^2)/2\rho^2$ and
$q=e^{-\kappa}$, the centre response of the truncated profile is
\begin{equation}
	R(\mathbf{0},\sigma)
	\;=\; \frac{A}{1-p}\left[
		\frac{2\bigl(1-(1+\kappa)q\bigr)}{(1+\rho^2)^2}
		\;-\; \frac{2(1-q)}{1+\rho^2}
		\;+\; \frac{c^2\,q}{\rho^2}
		\right],
	\label{eq:supp:truncresponse}
\end{equation}
which recovers the ideal Laplacian-of-Gaussian response
$-2A\sigma_0^2\sigma^2/(\sigma_0^2+\sigma^2)^2$ as $c\to\infty$. Its
extremum lies at $\beta(c)\,\sigma_0$, where $\beta(c)$ is the pedestal
scale bias of the main paper,
$\beta(c)=\arg\max_{\rho}|R(\mathbf{0},\rho\sigma_0)|$. Because $\beta$
depends only on $c$, the truncated response is still scale-covariant
with a peak magnitude independent of $\sigma_0$, so dividing each detected
scale by $\beta(c)$ recovers $\sigma_0$ exactly under this continuous
model.

\section{Descriptor Training: Reference and Observation Patches}
\label{sec:supp:training}

At test time a query patch cropped from an image is matched against a
descriptor computed from the reference pattern, so the training objective
must contain that pairing explicitly. It does. For every board, reference
patches are rendered from the reference pattern and the known blob geometry, and are
written into the same patch tensor as the real observations, keyed by the same
$(\text{board hash},\,\text{blob index})$ identity. A reference patch is
therefore an ordinary member of its supervised-contrastive class: it serves as
an anchor whose positives are the real observations of that blob, and as a
positive when a real observation is the anchor.

The batch sampler ensures that reference patches are available for real patches by concatenating the patches for all tracks sharing a blob id including the reference patch and treating them as a unit. Unmatched detections enter each batch as confusers with unique singleton labels: they never act as anchors and never supply positives, appearing only in other anchors' negative denominators.  A batch is made up of \(1/2\) continuous sequences, and \(1/2\) confusers.

\section{Detection, Matching and Verification Settings}
\label{sec:supp:settings}

\Cref{tab:supp:settings} lists every operational parameter of the pipeline as
run for the results in this paper. These are transcribed from the run
configuration shipped with the released dataset record, which each driver
copies beside its output, so a run's artifacts always carry the configuration
that produced them.

\begin{table}[!ht]
	\centering\small
	\begin{tabular}{@{}llp{0.42\linewidth}@{}}
		\toprule
		stage                    & parameter                       & value                                                                       \\
		\midrule
		detect                   & affine warps                    & IMAS-$25$: identity, plus tilt $2.62$ at $8$ angles $\pi k/8$ and tilt $5.18$ at $16$ angles $\pi k/16$ \\
		                         & pre-upsampling                  & $2\times$ (scale space starts at octave $-1$)                               \\
		                         & DoG peak threshold              & $0.01$                                                                      \\
		                         & DoG edge threshold              & $5.0$                                                                       \\
		                         & cross-warp NMS                  & elliptical, Mahalanobis cutoff $2.0$                                        \\
		                         & detections kept                 & unbounded (no response-rank cap)                                            \\
		\midrule
		adapt                    & scan window                     & $2.0\,\sigma$                                                               \\
		                         & max IRLS iterations             & $20$                                                                        \\
		                         & convergence tolerance           & $10^{-6}$ relative                                                          \\
		                         & storage / accumulation          & \texttt{Float32} / \texttt{Float64}                                         \\
		                         & validity guards                 & scale $\le0.1\sqrt{WH}$, scale ratio $\le4$, clip ratio $\le0.2$, aspect $\le10$ \\
		\midrule
		canonicalize             & log-polar grid                  & $64\times64$                                                                \\
		                         & inner radius                    & $c = 2.0$                                                            \\
		                         & outer radius $c_{\mathrm{out}}$ & $96$ (in units of $\sigma$)                                                                \\
		\midrule
		match                    & descriptor similarity threshold & $0.6$ (cosine)                                                              \\
		                         & first pass                      & mutual nearest neighbour                                                    \\
		                         & second pass                     & $k$-nearest neighbour, $k=3$                                                \\
		\midrule
		verify                   & geometric model                 & P3P (calibrated)                                                            \\
		                         & $N_{\mathrm{min}}$ inliers      & $9$ (inclusive)                                                             \\
		                         & inlier half-width $a$           & $5$ px                                                                      \\
		                         & RANSAC iterations               & $100$ minimum, $10\,000$ maximum                                            \\
		                         & RANSAC confidence               & $0.99$                                                                      \\
		                         & local optimization              & Student-$t$/uniform IRLS, seeded per board                                   \\
		                         & appearance certification        & $\ge50\%$ of inliers at cosine similarity $\ge0.6$                           \\
		                         & mirror branch test              & enabled (sign of the score difference only)                                 \\
		\bottomrule
	\end{tabular}
	\caption{Operational settings for the reported runs. The RANSAC seed is
		fixed and each board's stream is derived from it, so the fit stage is
		bit-reproducible given the extracted features.}
	\label{tab:supp:settings}
\end{table}

\section{Pipeline}
\label{sec:supp:pipeline}

The pipeline of the main paper in full, in the order the stages run.
Stages (1)--(4) are computed once per image and shared across all
candidate boards; (5)--(8) run per board and compete for the same
detections.

\begin{algorithm}[!ht]
	\caption{Detection, identification and pose for a gallery of boards.
		Stages (1)--(4) run once per image; (5)--(8) run per board and
		compete for the same detections, so committing a board claims its
		detections before the next fit.}
	\label{alg:pipeline}
	\begin{algorithmic}[1]
		\Require image $I$; gallery $\mathcal{G}$ of boards with reference
		descriptors and geometry $\mathcal{B}$; camera (optional)
		\Ensure accepted $(\text{board id},\,\text{pose})$ pairs
		\State $\mathcal{F}\gets$ \Call{DetectAdaptDescribe}{$I$}
		\Comment{(1)--(4), once per image}
		\State $C\gets\varnothing$
		\Comment{claimed detections}
		\ForAll{$(\text{strategy},s_{\min})\in\{(\text{mutual\,NN},\,s),\ (k\text{NN},\,s')\}$}
		\Comment{(5) strict, then permissive}
		\State $M\gets$ \Call{Match}{$\mathcal{G},\ \mathcal{F}\setminus C,\ \text{strategy},\ s_{\min}$}
		\While{$\mathcal{G}\neq\varnothing$}
		\State $A\gets\varnothing$
		\Comment{this round's admissible candidates}
		\ForAll{$g\in\mathcal{G}$}
		\State $M_g\gets$ \Call{ResolveOneToOne}{$M_g\setminus C$}
		\Comment{re-resolved against the current claims}
		\State \textbf{if} $|M_g|<N_{\min}$ \textbf{then continue}
		\Comment{match floor}
		\State $(\vtheta,I_g)\gets$ \Call{LoRansac}{$M_g$}
		\Comment{(6) spatial verification}
		\State $\vtheta\gets$ \Call{ResolveMirror}{$\vtheta,I_g$}
		\Comment{planar pose ambiguity}
		\State \textbf{if} $|I_g|<N_{\min}$ \textbf{then continue}
		\State \textbf{if} \Call{AppearanceSupport}{$I_g$} $<\tfrac12$ \textbf{then continue}
		\Comment{(7) certification}
		\State $A\gets A\cup\{(g,\vtheta,I_g)\}$
		\EndFor
		\State \textbf{if} $A=\varnothing$ \textbf{then break}
		\State $(g,\vtheta,I_g)\gets\arg\max_{A}\,(|I_g|,\ \text{score})$
		\Comment{commit the best-supported board}
		\State $C\gets C\cup\{$detections inside the quad of $g$ under $\vtheta\}$
		\Comment{(8) claim}
		\State accept $(g,\vtheta)$;\ \ $\mathcal{G}\gets\mathcal{G}\setminus\{g\}$
		\EndWhile
		\EndFor
		\State \Return accepted pairs
	\end{algorithmic}
\end{algorithm}

\section{Experimental Setup}
\label{sec:supp:setup}

\begin{figure}[!ht]
	\centering
	\begin{minipage}[c]{0.025\textwidth}
		\centering\rotatebox{90}{\small BlobBoards}
	\end{minipage}\hfill
	\begin{minipage}[c]{0.29\textwidth}
		\includegraphics[width=\textwidth]{arxiv/images/experiment/bb/DSC04563_val.png}
	\end{minipage}\hfill
	\begin{minipage}[c]{0.29\textwidth}
		\includegraphics[width=\textwidth]{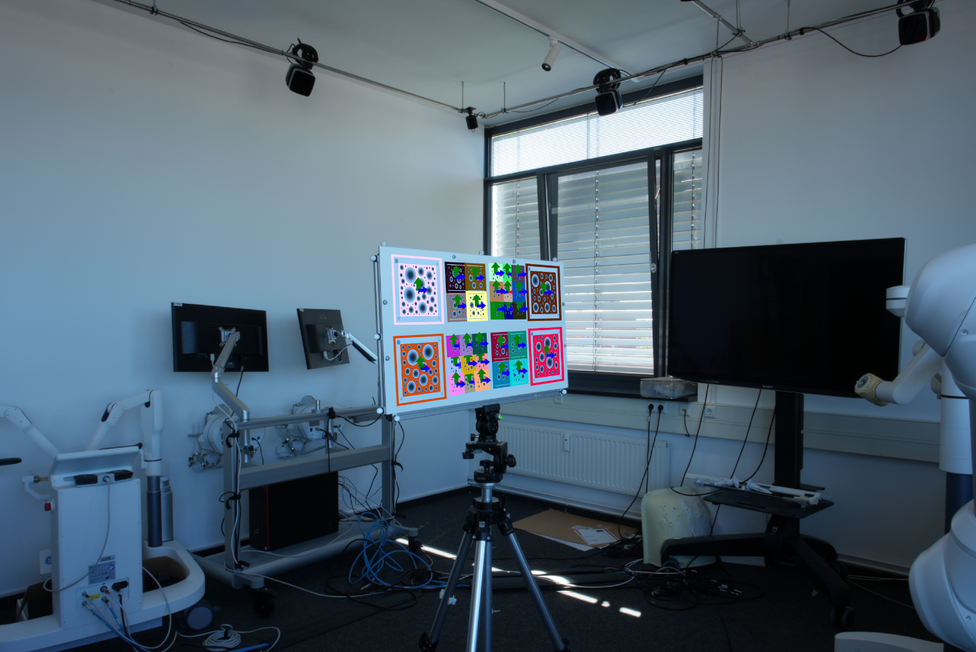}
	\end{minipage}\hfill
	\begin{minipage}[c]{0.29\textwidth}
		\includegraphics[width=\textwidth]{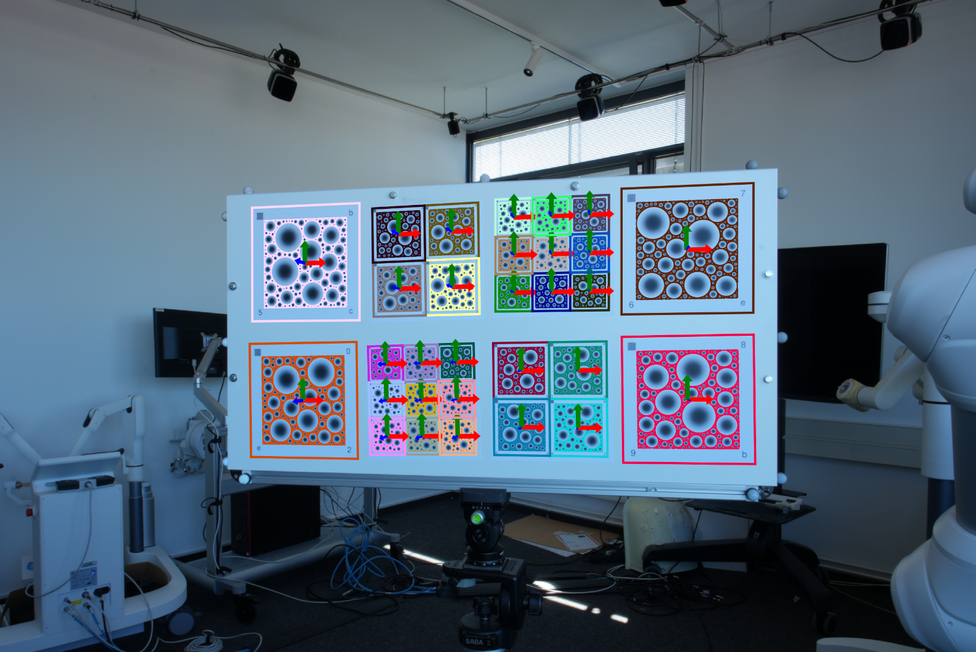}
	\end{minipage}

	\vspace{0.4em}

	\begin{minipage}[c]{0.025\textwidth}
		\centering\rotatebox{90}{\small AprilTag}
	\end{minipage}\hfill
	\begin{minipage}[c]{0.29\textwidth}
		\includegraphics[width=\textwidth]{arxiv/images/experiment/apr/DSC04465_val.png}
	\end{minipage}\hfill
	\begin{minipage}[c]{0.29\textwidth}
		\includegraphics[width=\textwidth]{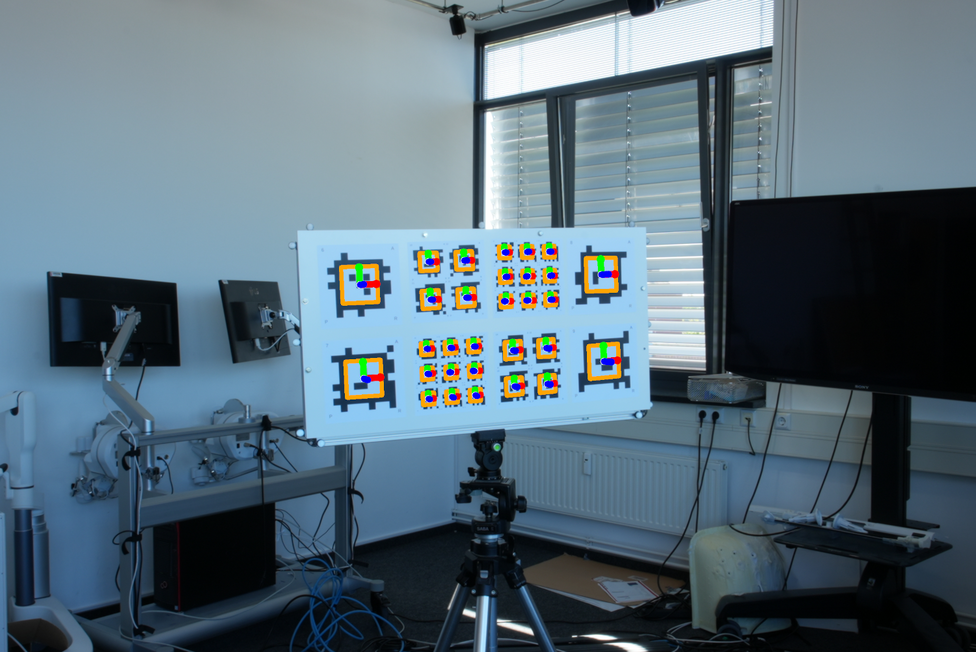}
	\end{minipage}\hfill
	\begin{minipage}[c]{0.29\textwidth}
		\includegraphics[width=\textwidth]{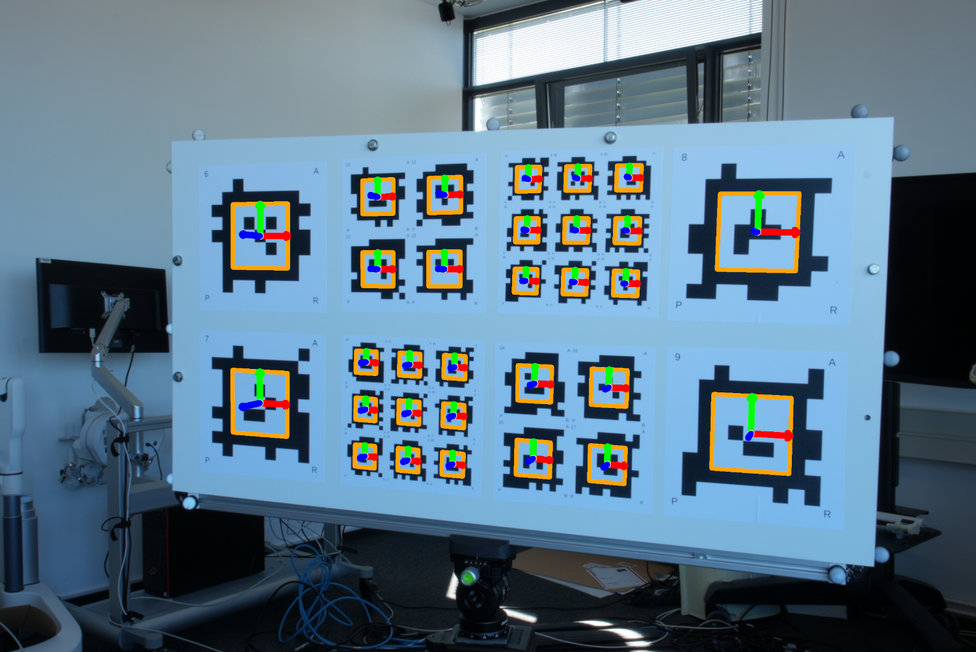}
	\end{minipage}

	\vspace{0.4em}

	\begin{minipage}[c]{0.025\textwidth}
		\centering\rotatebox{90}{\small ArUco}
	\end{minipage}\hfill
	\begin{minipage}[c]{0.29\textwidth}
		\includegraphics[width=\textwidth]{arxiv/images/experiment/aru/DSC04375_val.png}
	\end{minipage}\hfill
	\begin{minipage}[c]{0.29\textwidth}
		\includegraphics[width=\textwidth]{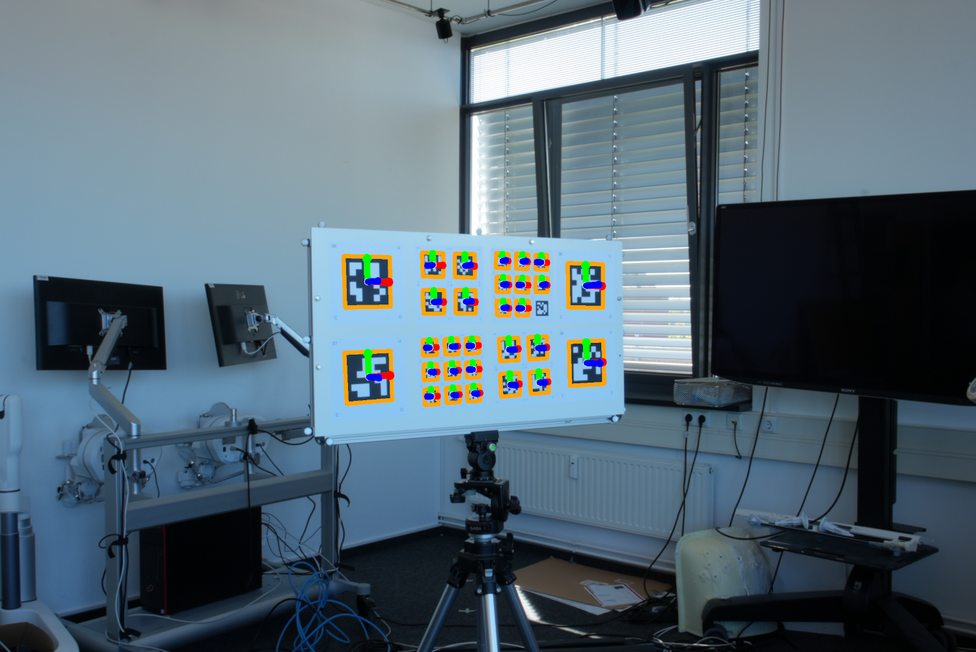}
	\end{minipage}\hfill
	\begin{minipage}[c]{0.29\textwidth}
		\includegraphics[width=\textwidth]{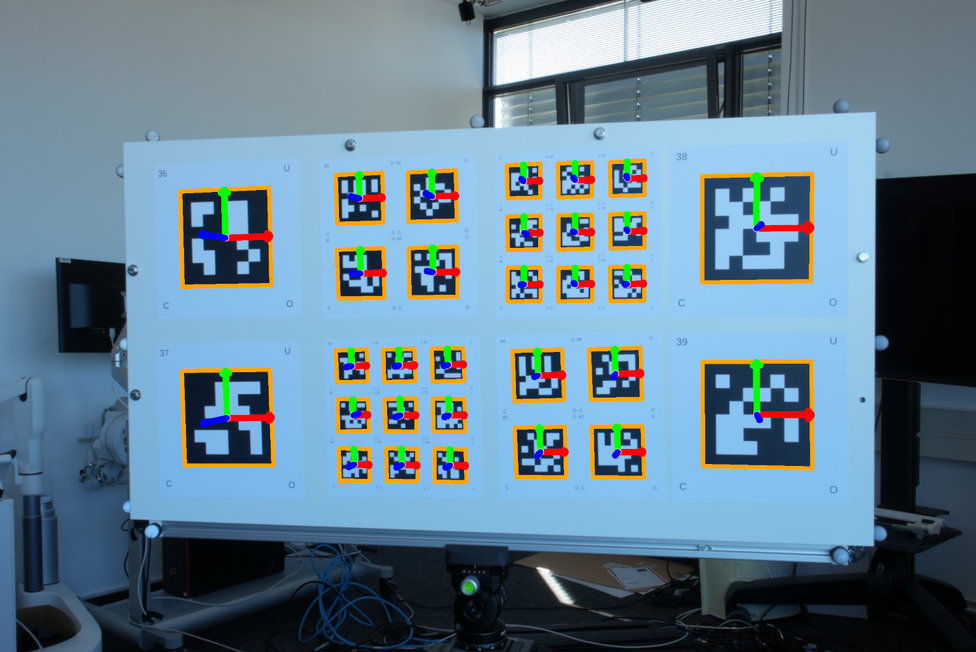}
	\end{minipage}

	\vspace{0.4em}

	\caption{Experimental setup. BlobBoards, AprilTag, and ArUco are mounted
		on matched rigid panels containing $30$ markers at three physical
		scales: $18$ small, $8$ medium, and $4$ large. The panel is observed
		at varying distances and obliquities while its $6$-DoF pose is tracked
		by motion capture. Estimated board coordinate frames are overlaid; for
		BlobBoards, the reference pattern is additionally projected under the
		recovered pose.}
	\label{fig:supp:setup}
\end{figure}

\Cref{fig:supp:setup} shows three further captures per marker type, beyond
the single capture per type in the main paper. All three panels
carry $30$ markers at three physical scales and are observed from the same
stations, so the rows are directly comparable; estimated board frames are
overlaid, and for BlobBoards the reference pattern is additionally projected
under the recovered pose.

\section{Hand--Eye Calibration Diagnostics}
\label{sec:supp:handeye}

\begin{table}[!ht]
	\centering\small
\begin{tabular}{@{}lrrrr@{}}
\toprule
system & frames & recovered & resid.\ mm & resid.\ deg \\
\midrule
BlobBoards & $41$ & $37$ & $0.68$ & $0.270$ \\
AprilTag & $22$ & $0$ & $2.54$ & $0.432$ \\
ArUco & $15$ & $0$ & $2.81$ & $1.885$ \\
\bottomrule
\end{tabular}

	\caption{Hand--eye calibration diagnostics, one fit per marker system over all $\bbHeMarkers$ boards. \emph{Residual} is the noise scale of the pose residual the staged linear fit reports, in millimetres and degrees; \emph{recovered} counts calibration frames whose untracked camera pose was substituted. The camera is static; its rig was untracked in $\bbHeRecovered$ of the $\bbHeFrames$ BlobBoard calibration frames, which use a substituted constant camera pose admitted only under an explicit constancy gate on the motion-capture reconstruction and the image background.}
	\label{tab:handeye}
\end{table}

\noindent\textbf{Independence of the ground truth.}~ Exactly two observation
streams enter the calibration: the motion-capture rig poses $\mathbf{A}_i$
and the board poses $\hat{\mathbf{B}}_{im}$ reported by the marker system
under test (notation of the main paper). Three properties bound what the
fitted mounts can absorb. First, an explicit calibration set is used,
disjoint from every evaluation frame ($41$, $22$ and $15$ usable calibration
frames for BlobBoards, AprilTag and ArUco, against $56$, $57$ and $57$
evaluation frames), so no evaluation frame contributes to the transforms it
is later scored against. Second, the calibration is fitted once per marker
system, from that system's own detections, and then held fixed: every system
is scored against ground truth from its own detector and never a
competitor's, and since the three panels are physically distinct objects a
shared set of mounts does not exist in any case. Third, the fit is heavily
overdetermined and spans the same distances and obliquities as the
evaluation (\cref{sec:supp:derivations}), so a rigid mount can absorb only
an error that is constant over that span. A constant per-system bias is
absorbed for each system alike; scale-, distance- and obliquity-dependent
degradation survives into the numbers the main paper reports. The residual noise scale of the BlobBoard fit is $\bbHeResidTransMM$\,mm and $\bbHeResidRotDeg^\circ$, well below the pose differences the evaluation must resolve (\cref{tab:handeye}).

\section{Hand--Eye Estimator}
\label{sec:supp:derivations}

\noindent\textbf{Conditioning of the fit.}~ The calibration is heavily overdetermined: up to $30$ boards observed in each
of $\ucoHeFrames$--$\bbHeFrames$ calibration frames constrain $6+6\times30$ parameters, and each
board is seen over the same span of distances and obliquities as in the
evaluation. Only a detector error that is constant over that span is
indistinguishable from a rigid mount error and therefore absorbable; any
scale-, distance- or obliquity-dependent component is not.

\noindent\textbf{Staged hand--eye solve.}~ Each of the three stages of the main
paper is in closed form: the shared rotation of $\mathbf{X}$ by Kabsch
alignment of the rotation axes of the relative motions pooled over all
markers, each mount rotation as the chordal mean of its per-frame estimates,
and all translations by one joint least-squares solve. The estimator is
deliberately linear: a pose-residual nonlinear refinement was implemented and
discarded, because it reduced in-sample cost by the amount its own quadratic
model predicts yet extrapolated worse across the evaluation range, the
signature of model misspecification absorbing an unmodelled systematic along a
weakly constrained direction.

\section{Pooled Error Distributions}
\label{sec:supp:pooled}

The main paper breaks recall down by marker size, because that is where the
methods separate. Pooling the three sizes gives the aggregate view below, in
which the tag curves flatten early where the two-fold planar pose ambiguity
takes over.

\begin{figure}[!t]
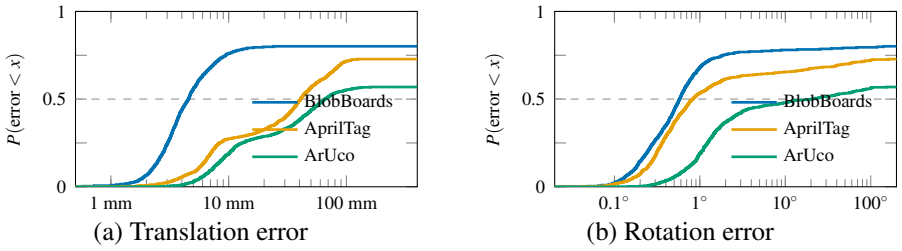

	\centering
\centering
\begin{tabular}{@{}cc@{}}
\input{figures/floorplan/hist_trans} & \input{figures/floorplan/hist_rot} \\
(a) Translation error & (b) Rotation error \\
\end{tabular}

	\caption{Recall as a function of the translation and rotation error
		threshold, pooled over all marker sizes. Every visible board
		constitutes an attempt; a missed detection never enters the numerator,
		so each curve saturates at the corresponding probability of detection.
		Dashed lines mark recall $0.5$.}
	\label{fig:supp:pod_trans}
\end{figure}

\section{Unoccluded Performance by Angle and Distance}
\label{sec:supp:unoccluded}

The floor-plan figure of the main paper reports the unoccluded result over the capture floor
plan, with distance and viewing angle on the two axes of a single plan and the
per-station median as colour. The figure below is the same measurements viewed
as marginals: each quantity against obliquity, and translation additionally
against camera distance. It adds what a per-station median colour cannot show
--- the interquartile band, and so the spread of the error at each station, not
only its centre.

\begin{figure}[!t]
	\centering
\centering
\setlength{\tabcolsep}{2pt}
\providecolor{bbGreen}{HTML}{009E73}
\providecolor{bbBlue}{HTML}{0072B2}
\providecolor{bbOrange}{HTML}{E69F00}
\begin{tabular}{@{}cc@{}}
\providecolor{bbGreen}{HTML}{009E73}
\providecolor{bbBlue}{HTML}{0072B2}
\providecolor{bbOrange}{HTML}{E69F00}
\begin{tikzpicture}[baseline=(current bounding box.center)]
\begin{axis}[width=5.6cm, height=4.0cm,
  xlabel={obliquity ($^\circ$)}, ylabel={rotation error ($^\circ$)},
  ymode=log, log ticks with fixed point, xtick={-90,-60,-30,0,30,60,90},
             ytick={0.1,0.5,1,5,10},
  label style={font=\scriptsize}, tick label style={font=\scriptsize},
  ]
\addplot[draw=none, fill=bbBlue, fill opacity=0.15, forget plot] coordinates {(-80,0.3715) (-70,0.3137) (-60,0.3876) (-50,0.4385) (-40,0.4881) (-30,0.5188) (-20,0.5521) (-10,0.7286) (0,0.8319) (10,0.3568) (20,0.1661) (30,0.2104) (40,0.1359) (50,0.1306) (60,0.1603) (70,0.1479) (80,0.2295) (80,0.2638) (70,0.5411) (60,0.2879) (50,0.2433) (40,0.2607) (30,0.291) (20,0.3658) (10,1.122) (0,2.876) (-10,1.968) (-20,1.125) (-30,0.8604) (-40,0.8272) (-50,0.6867) (-60,0.7185) (-70,0.5926) (-80,0.5723)} \closedcycle;
\addplot[draw=bbBlue, line width=1.0pt, mark=*, mark size=1.1pt, mark options={fill=bbBlue, draw=bbBlue}] coordinates {(-80,0.4544) (-70,0.4474) (-60,0.5253) (-50,0.5524) (-40,0.6342) (-30,0.677) (-20,0.7455) (-10,1.246) (0,1.446) (10,0.6195) (20,0.2642) (30,0.2485) (40,0.1951) (50,0.1798) (60,0.2159) (70,0.2991) (80,0.2542)};
\addplot[draw=none, fill=bbOrange, fill opacity=0.15, forget plot] coordinates {(-80,0.377) (-70,0.3623) (-60,0.3232) (-50,0.312) (-40,0.2584) (-30,0.3126) (-20,0.3196) (-10,0.7579) (0,1.121) (10,0.6806) (20,0.3676) (30,0.2502) (40,0.2097) (50,0.2036) (60,0.2018) (70,0.2731) (80,0.6432) (80,0.814) (70,0.4615) (60,0.4774) (50,0.5855) (40,0.5672) (30,0.8289) (20,1.22) (10,3.012) (0,7.141) (-10,5.246) (-20,1.305) (-30,0.8632) (-40,0.4717) (-50,0.5743) (-60,0.5639) (-70,0.6674) (-80,0.5496)} \closedcycle;
\addplot[draw=bbOrange, line width=1.0pt, mark=*, mark size=1.1pt, mark options={fill=bbOrange, draw=bbOrange}] coordinates {(-80,0.4934) (-70,0.4571) (-60,0.4584) (-50,0.4092) (-40,0.3383) (-30,0.4399) (-20,0.6094) (-10,1.549) (0,2.079) (10,1.399) (20,0.6406) (30,0.399) (40,0.3714) (50,0.3226) (60,0.3123) (70,0.3232) (80,0.756)};
\addplot[draw=none, fill=bbGreen, fill opacity=0.15, forget plot] coordinates {(-80,1.05) (-70,0.5381) (-60,0.9238) (-50,0.8018) (-40,0.9087) (-30,1.014) (-20,1.008) (-10,1.793) (0,1.63) (10,1.505) (20,1.106) (30,0.7778) (40,0.4942) (50,0.5263) (60,0.6063) (70,0.6953) (80,0.473) (80,1.004) (70,1.165) (60,1.109) (50,1.762) (40,1.89) (30,2.123) (20,4.102) (10,12.16) (0,10.33) (-10,13.74) (-20,3.959) (-30,3.37) (-40,1.892) (-50,1.735) (-60,1.507) (-70,1.247) (-80,1.505)} \closedcycle;
\addplot[draw=bbGreen, line width=1.0pt, mark=*, mark size=1.1pt, mark options={fill=bbGreen, draw=bbGreen}] coordinates {(-80,1.24) (-70,0.9464) (-60,1.245) (-50,1.055) (-40,1.432) (-30,1.636) (-20,1.762) (-10,3.451) (0,6.835) (10,3.203) (20,2.113) (30,1.158) (40,1.089) (50,0.8569) (60,0.8674) (70,0.9434) (80,0.9313)};
\end{axis}
\end{tikzpicture} & 
\providecolor{bbGreen}{HTML}{009E73}
\providecolor{bbBlue}{HTML}{0072B2}
\providecolor{bbOrange}{HTML}{E69F00}
\begin{tikzpicture}[baseline=(current bounding box.center)]
\begin{axis}[width=5.6cm, height=4.0cm,
  xlabel={obliquity ($^\circ$)}, ylabel={translation error (mm)},
  ymode=log, log ticks with fixed point, xtick={-90,-60,-30,0,30,60,90},
             ytick={2,5,10,25,50,100},
  label style={font=\scriptsize}, tick label style={font=\scriptsize},
  ]
\addplot[draw=none, fill=bbBlue, fill opacity=0.15, forget plot] coordinates {(-80,2.322) (-70,4.679) (-60,3.401) (-50,3.625) (-40,3.233) (-30,2.919) (-20,2.755) (-10,1.953) (0,2.685) (10,2.497) (20,2.155) (30,2.439) (40,3.033) (50,2.193) (60,3.572) (70,3.774) (80,2.418) (80,3.07) (70,5.728) (60,5.597) (50,6.376) (40,5.333) (30,4.681) (20,5.011) (10,4.488) (0,4.056) (-10,5.445) (-20,5.202) (-30,5.302) (-40,5.47) (-50,6.071) (-60,6.76) (-70,6.246) (-80,4.779)} \closedcycle;
\addplot[draw=bbBlue, line width=1.0pt, mark=*, mark size=1.1pt, mark options={fill=bbBlue, draw=bbBlue}] coordinates {(-80,3.217) (-70,5.214) (-60,4.196) (-50,4.6) (-40,3.874) (-30,3.69) (-20,3.333) (-10,2.361) (0,3.032) (10,3.252) (20,3.398) (30,3.055) (40,3.812) (50,3.552) (60,4.355) (70,4.578) (80,2.784)};
\addplot[draw=none, fill=bbOrange, fill opacity=0.15, forget plot] coordinates {(-80,5.085) (-70,4.531) (-60,6.735) (-50,7.003) (-40,7.791) (-30,7.062) (-20,5.566) (-10,5.308) (0,7.082) (10,14.86) (20,15.05) (30,7.369) (40,7.324) (50,7.967) (60,6.444) (70,6.322) (80,3.268) (80,6.942) (70,12.6) (60,25.65) (50,32.54) (40,41.9) (30,52.3) (20,49.13) (10,51.09) (0,66.4) (-10,77.11) (-20,73.89) (-30,70.59) (-40,66.0) (-50,44.63) (-60,37.26) (-70,16.73) (-80,6.446)} \closedcycle;
\addplot[draw=bbOrange, line width=1.0pt, mark=*, mark size=1.1pt, mark options={fill=bbOrange, draw=bbOrange}] coordinates {(-80,5.921) (-70,5.947) (-60,18.77) (-50,30.6) (-40,37.88) (-30,38.81) (-20,23.98) (-10,21.71) (0,38.5) (10,36.9) (20,35.1) (30,27.27) (40,25.5) (50,19.19) (60,14.23) (70,9.145) (80,4.838)};
\addplot[draw=none, fill=bbGreen, fill opacity=0.15, forget plot] coordinates {(-80,7.434) (-70,9.187) (-60,7.717) (-50,6.762) (-40,8.514) (-30,7.284) (-20,8.44) (-10,8.807) (0,8.897) (10,8.47) (20,10.3) (30,9.932) (40,9.752) (50,8.755) (60,5.875) (70,8.062) (80,8.922) (80,21.74) (70,29.12) (60,26.91) (50,36.13) (40,55.46) (30,43.81) (20,54.38) (10,56.18) (0,49.19) (-10,58.85) (-20,60.0) (-30,56.77) (-40,49.26) (-50,41.11) (-60,34.34) (-70,33.62) (-80,12.0)} \closedcycle;
\addplot[draw=bbGreen, line width=1.0pt, mark=*, mark size=1.1pt, mark options={fill=bbGreen, draw=bbGreen}] coordinates {(-80,10.06) (-70,11.58) (-60,11.8) (-50,13.63) (-40,27.33) (-30,30.84) (-20,39.52) (-10,27.27) (0,25.3) (10,28.58) (20,36.64) (30,30.91) (40,31.3) (50,14.54) (60,10.09) (70,12.19) (80,13.07)};
\end{axis}
\end{tikzpicture} \\
(a) Rotation error & (b) Translation error \\[6pt]
\providecolor{bbGreen}{HTML}{009E73}
\providecolor{bbBlue}{HTML}{0072B2}
\providecolor{bbOrange}{HTML}{E69F00}
\begin{tikzpicture}[baseline=(current bounding box.center)]
\begin{axis}[width=5.6cm, height=4.0cm,
  xlabel={camera distance}, ylabel={translation error (mm)},
  ymode=log, log ticks with fixed point, ytick={2,5,10,25,50,100},
             xtick={1000,2000,3000}, xticklabels={{1\,m},{2\,m},{3\,m}},
  label style={font=\scriptsize}, tick label style={font=\scriptsize},
  ]
\addplot[draw=none, fill=bbBlue, fill opacity=0.15, forget plot] coordinates {(1000,2.474) (2000,3.124) (3000,3.561) (3000,8.377) (2000,5.297) (1000,3.722)} \closedcycle;
\addplot[draw=bbBlue, line width=1.0pt, mark=*, mark size=1.1pt, mark options={fill=bbBlue, draw=bbBlue}] coordinates {(1000,3.039) (2000,4.075) (3000,5.947)};
\addplot[draw=none, fill=bbOrange, fill opacity=0.15, forget plot] coordinates {(1000,4.398) (2000,21.48) (3000,43.8) (3000,81.55) (2000,40.72) (1000,7.29)} \closedcycle;
\addplot[draw=bbOrange, line width=1.0pt, mark=*, mark size=1.1pt, mark options={fill=bbOrange, draw=bbOrange}] coordinates {(1000,6.156) (2000,31.89) (3000,66.29)};
\addplot[draw=none, fill=bbGreen, fill opacity=0.15, forget plot] coordinates {(1000,6.328) (2000,30.35) (3000,48.29) (3000,93.06) (2000,55.98) (1000,10.64)} \closedcycle;
\addplot[draw=bbGreen, line width=1.0pt, mark=*, mark size=1.1pt, mark options={fill=bbGreen, draw=bbGreen}] coordinates {(1000,8.277) (2000,41.25) (3000,65.38)};
\end{axis}
\end{tikzpicture} & 
\providecolor{bbGreen}{HTML}{009E73}
\providecolor{bbBlue}{HTML}{0072B2}
\providecolor{bbOrange}{HTML}{E69F00}
\begin{tikzpicture}[baseline=(current bounding box.center)]
\begin{axis}[width=5.6cm, height=4.0cm,
  xlabel={obliquity ($^\circ$)}, ylabel={$P(\mathrm{detection})$},
  ymin=0, ymax=1.05, ytick={0,0.25,0.5,0.75,1}, yticklabels={0,,0.5,,1},
             xtick={-90,-60,-30,0,30,60,90},
  label style={font=\scriptsize}, tick label style={font=\scriptsize},
  ]
\addplot[draw=bbBlue, line width=1.0pt, mark=*, mark size=1.1pt, mark options={fill=bbBlue, draw=bbBlue}] coordinates {(-90,0.0) (-80,0.3556) (-70,0.8) (-60,1.0) (-50,1.0) (-40,1.0) (-30,1.0) (-20,1.0) (-10,1.0) (0,1.0) (10,1.0) (20,1.0) (30,1.0) (40,1.0) (50,1.0) (60,0.9667) (70,0.7444) (80,0.08889) (90,0.0)};
\addplot[draw=bbOrange, line width=1.0pt, mark=*, mark size=1.1pt, mark options={fill=bbOrange, draw=bbOrange}] coordinates {(-90,0.0) (-80,0.06667) (-70,0.4889) (-60,0.7556) (-50,0.9111) (-40,1.0) (-30,1.0) (-20,1.0) (-10,1.0) (0,1.0) (10,1.0) (20,1.0) (30,1.0) (40,1.0) (50,0.9889) (60,0.8667) (70,0.5333) (80,0.2222) (90,0.0)};
\addplot[draw=bbGreen, line width=1.0pt, mark=*, mark size=1.1pt, mark options={fill=bbGreen, draw=bbGreen}] coordinates {(-90,0.0) (-80,0.2667) (-70,0.5111) (-60,0.5111) (-50,0.6333) (-40,0.7889) (-30,0.7889) (-20,0.8) (-10,0.7889) (0,0.8) (10,0.8) (20,0.8) (30,0.8) (40,0.8) (50,0.6556) (60,0.5111) (70,0.5) (80,0.05556) (90,0.0)};
\end{axis}
\end{tikzpicture} \\
(c) Translation error by distance & (d) Probability of detection \\
\end{tabular}\\[4pt]
\begin{tikzpicture}[baseline]
\draw[bbBlue, line width=1.2pt] (0.0cm,0) -- (0.6cm,0);
\node[anchor=west, font=\small] at (0.7cm,0) {BlobBoards};
\draw[bbOrange, line width=1.2pt] (3.0cm,0) -- (3.6cm,0);
\node[anchor=west, font=\small] at (3.7cm,0) {AprilTag};
\draw[bbGreen, line width=1.2pt] (6.0cm,0) -- (6.6cm,0);
\node[anchor=west, font=\small] at (6.7cm,0) {ArUco};
\end{tikzpicture}

	\caption{Unoccluded performance binned by obliquity and camera distance.
		Curves are per-bin medians, bands the interquartile range. Bins with
		fewer than five detections are omitted.}
	\label{fig:supp:unoccluded}
\end{figure}
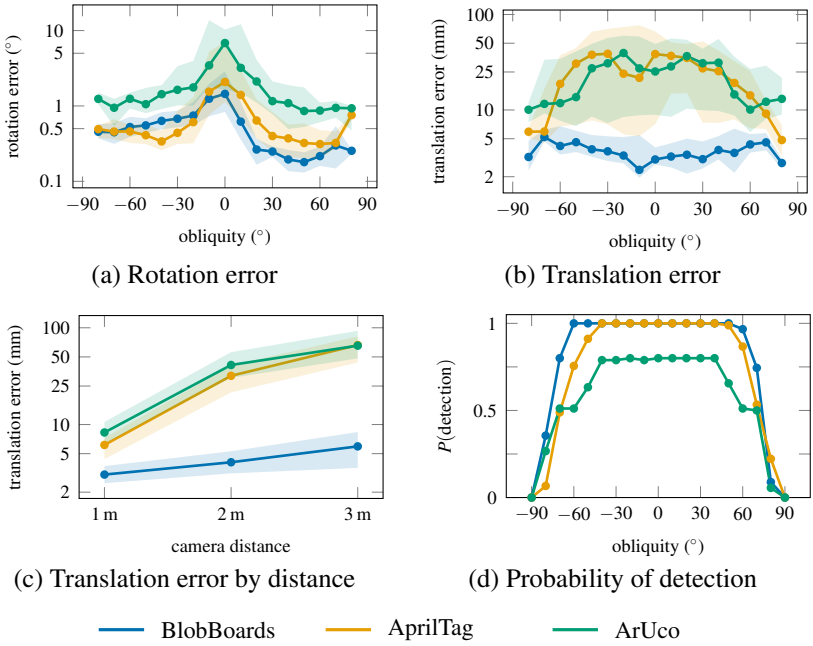

\section{Occlusion Examples}
\label{sec:supp:occexamples}

The synthetic occluder is one deterministic region per marker, constructed in
the pattern's own coordinate frame and warped into the image through the
ground-truth pose, so the same area fraction $f$ is hidden on every marker type
at every scale and viewpoint. Its boundary is a single $45^\circ$ line anchored
at the lower-left corner of the pattern. For $f\le\tfrac12$ the masked region
is the right triangle with legs $\sqrt{2f}$ of the pattern side; a triangle
cannot exceed half of a square, so for $f>\tfrac12$ the masked region is
instead the complement of the right triangle of area $(1-f)$ at the opposite
corner. Either way the boundary crosses two edges of the marker, which is what
breaks a quad-based detector. The realized covered fraction matches the target
exactly at every level of the sweep. We use this one placement rather than
averaging over occluder positions.

\section{Occlusion by Floor-Plan Level}
\label{sec:supp:occlevels}

The main paper quotes the $50\%$ level in prose and plots the sweep over
all ten levels. The plans below repeat the unoccluded floor plan's
encoding at $10\%$, $50\%$ and $70\%$, so the loss of coverage can be read
per station rather than only in aggregate. The $10\%$ level locates where
the two designs part: an occluder crossing the marker border breaks the
single quad AprilTag and ArUco depend on, while BlobBoards lose only the
blobs the occluder covers. By $50\%$ neither tag baseline returns poses
anywhere on the floor, and BlobBoard coverage is not confined to the near,
frontal stations where the boards are largest.

\begin{figure*}[h]
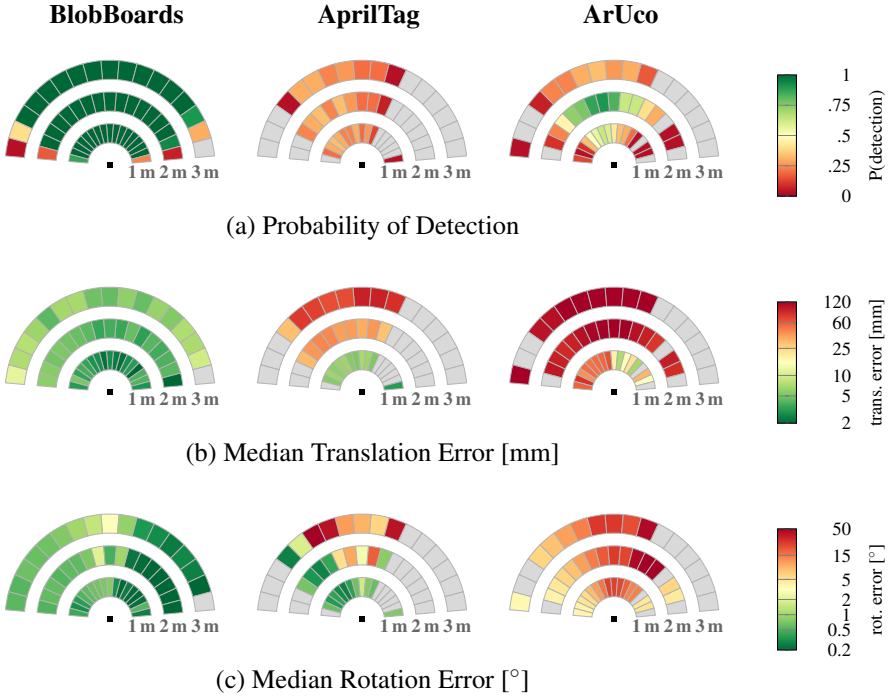

	\centering
\providecommand{\fpscale}{1}\renewcommand{\fpscale}{0.42}   
\setlength{\tabcolsep}{3pt}
\centering
\begin{tabular}{@{}ccc l@{}}
\textbf{BlobBoards} & \textbf{AprilTag} & \textbf{ArUco} & \\[2pt]
\input{figures/floorplan_occ_10/floorplan_blob_det} & \input{figures/floorplan_occ_10/floorplan_april_det} & \input{figures/floorplan_occ_10/floorplan_aruco_det} & 
\begin{tikzpicture}[baseline=(current bounding box.north)]
\pgfplotsset{colormap={rdylgnrev}{rgb=(0.0,0.408,0.216) rgb=(0.0729,0.5423,0.286) rgb=(0.2297,0.6581,0.3457) rgb=(0.4359,0.7567,0.392) rgb=(0.6151,0.8353,0.412) rgb=(0.7653,0.9001,0.4897) rgb=(0.8936,0.955,0.6033) rgb=(1.0,1.0,0.749) rgb=(0.9971,0.9129,0.6033) rgb=(0.9943,0.794,0.4743) rgb=(0.987,0.6456,0.3633) rgb=(0.962,0.4634,0.2797) rgb=(0.8919,0.2904,0.2001) rgb=(0.787,0.1343,0.1519) rgb=(0.647,0.0,0.149)}, colormap={rdylgn}{rgb=(0.647,0.0,0.149) rgb=(0.787,0.1343,0.1519) rgb=(0.8919,0.2904,0.2001) rgb=(0.962,0.4634,0.2797) rgb=(0.987,0.6456,0.3633) rgb=(0.9943,0.794,0.4743) rgb=(0.9971,0.9129,0.6033) rgb=(1.0,1.0,0.749) rgb=(0.8936,0.955,0.6033) rgb=(0.7653,0.9001,0.4897) rgb=(0.6151,0.8353,0.412) rgb=(0.4359,0.7567,0.392) rgb=(0.2297,0.6581,0.3457) rgb=(0.0729,0.5423,0.286) rgb=(0.0,0.408,0.216)}}
\begin{axis}[hide axis, scale only axis, width=0pt, height=1.6cm,
  colormap name=rdylgn, colorbar, point meta min=0.0, point meta max=1.0,
  colorbar style={height=1.6cm, width=0.25cm, ylabel={P(detection)}, ylabel style={font=\scriptsize},
    ytick={0.0,0.25,0.5,0.75,1.0}, yticklabels={0,.25,.5,.75,1}, tick label style={font=\scriptsize},
    yticklabel style={text width=1.7em, align=right}}]
  \addplot[draw=none] coordinates {(0,0)};
\end{axis}
\end{tikzpicture} \\[2pt]
\multicolumn{3}{c}{(a) Probability of Detection} & \\[8pt]
\input{figures/floorplan_occ_10/floorplan_blob_trans} & \input{figures/floorplan_occ_10/floorplan_april_trans} & \input{figures/floorplan_occ_10/floorplan_aruco_trans} & 
\begin{tikzpicture}[baseline=(current bounding box.north)]
\pgfplotsset{colormap={rdylgnrev}{rgb=(0.0,0.408,0.216) rgb=(0.0729,0.5423,0.286) rgb=(0.2297,0.6581,0.3457) rgb=(0.4359,0.7567,0.392) rgb=(0.6151,0.8353,0.412) rgb=(0.7653,0.9001,0.4897) rgb=(0.8936,0.955,0.6033) rgb=(1.0,1.0,0.749) rgb=(0.9971,0.9129,0.6033) rgb=(0.9943,0.794,0.4743) rgb=(0.987,0.6456,0.3633) rgb=(0.962,0.4634,0.2797) rgb=(0.8919,0.2904,0.2001) rgb=(0.787,0.1343,0.1519) rgb=(0.647,0.0,0.149)}, colormap={rdylgn}{rgb=(0.647,0.0,0.149) rgb=(0.787,0.1343,0.1519) rgb=(0.8919,0.2904,0.2001) rgb=(0.962,0.4634,0.2797) rgb=(0.987,0.6456,0.3633) rgb=(0.9943,0.794,0.4743) rgb=(0.9971,0.9129,0.6033) rgb=(1.0,1.0,0.749) rgb=(0.8936,0.955,0.6033) rgb=(0.7653,0.9001,0.4897) rgb=(0.6151,0.8353,0.412) rgb=(0.4359,0.7567,0.392) rgb=(0.2297,0.6581,0.3457) rgb=(0.0729,0.5423,0.286) rgb=(0.0,0.408,0.216)}}
\begin{axis}[hide axis, scale only axis, width=0pt, height=1.6cm,
  colormap name=rdylgnrev, colorbar, point meta min=0.301, point meta max=2.0792,
  colorbar style={height=1.6cm, width=0.25cm, ylabel={trans.~error [mm]}, ylabel style={font=\scriptsize},
    ytick={0.301,0.699,1.0,1.3979,1.7782,2.0792}, yticklabels={2,5,10,25,60,120}, tick label style={font=\scriptsize},
    yticklabel style={text width=1.7em, align=right}}]
  \addplot[draw=none] coordinates {(0,0)};
\end{axis}
\end{tikzpicture} \\[2pt]
\multicolumn{3}{c}{(b) Median Translation Error [mm]} & \\[8pt]
\input{figures/floorplan_occ_10/floorplan_blob_rot} & \input{figures/floorplan_occ_10/floorplan_april_rot} & \input{figures/floorplan_occ_10/floorplan_aruco_rot} & 
\begin{tikzpicture}[baseline=(current bounding box.north)]
\pgfplotsset{colormap={rdylgnrev}{rgb=(0.0,0.408,0.216) rgb=(0.0729,0.5423,0.286) rgb=(0.2297,0.6581,0.3457) rgb=(0.4359,0.7567,0.392) rgb=(0.6151,0.8353,0.412) rgb=(0.7653,0.9001,0.4897) rgb=(0.8936,0.955,0.6033) rgb=(1.0,1.0,0.749) rgb=(0.9971,0.9129,0.6033) rgb=(0.9943,0.794,0.4743) rgb=(0.987,0.6456,0.3633) rgb=(0.962,0.4634,0.2797) rgb=(0.8919,0.2904,0.2001) rgb=(0.787,0.1343,0.1519) rgb=(0.647,0.0,0.149)}, colormap={rdylgn}{rgb=(0.647,0.0,0.149) rgb=(0.787,0.1343,0.1519) rgb=(0.8919,0.2904,0.2001) rgb=(0.962,0.4634,0.2797) rgb=(0.987,0.6456,0.3633) rgb=(0.9943,0.794,0.4743) rgb=(0.9971,0.9129,0.6033) rgb=(1.0,1.0,0.749) rgb=(0.8936,0.955,0.6033) rgb=(0.7653,0.9001,0.4897) rgb=(0.6151,0.8353,0.412) rgb=(0.4359,0.7567,0.392) rgb=(0.2297,0.6581,0.3457) rgb=(0.0729,0.5423,0.286) rgb=(0.0,0.408,0.216)}}
\begin{axis}[hide axis, scale only axis, width=0pt, height=1.6cm,
  colormap name=rdylgnrev, colorbar, point meta min=-0.699, point meta max=1.699,
  colorbar style={height=1.6cm, width=0.25cm, ylabel={rot.~error [$^\circ$]}, ylabel style={font=\scriptsize},
    ytick={-0.699,-0.301,0.0,0.301,0.699,1.1761,1.699}, yticklabels={0.2,0.5,1,2,5,15,50}, tick label style={font=\scriptsize},
    yticklabel style={text width=1.7em, align=right}}]
  \addplot[draw=none] coordinates {(0,0)};
\end{axis}
\end{tikzpicture} \\[2pt]
\multicolumn{3}{c}{(c) Median Rotation Error [$^\circ$]} & \\
\end{tabular}

	\caption{Pose accuracy over the capture floor plan at $10\%$ occlusion, in
		the same layout and encoding as the floor-plan figure of the main paper.
		Rings are tripod distance, wedges are $10^\circ$ rig stations; colour is
		(a) the fraction of the $30$ boards detected, and the per-station
		median over detected boards of (b) translation and (c) rotation error.
		\textbf{Gray wedges are stations with zero detections.} At this level
		the AprilTag and ArUco plans are already largely gray while the
		BlobBoards plan is intact --- the separation between the two designs is
		established well before the heavier levels shown below.}
	\label{fig:supp:occlusion10}
\end{figure*}

\begin{figure*}[h]
	\centering
\providecommand{\fpscale}{1}\renewcommand{\fpscale}{0.42}   
\setlength{\tabcolsep}{3pt}
\centering
\begin{tabular}{@{}ccc l@{}}
\textbf{BlobBoards} & \textbf{AprilTag} & \textbf{ArUco} & \\[2pt]
\input{figures/floorplan_occ_50/floorplan_blob_det} & \input{figures/floorplan_occ_50/floorplan_april_det} & \input{figures/floorplan_occ_50/floorplan_aruco_det} & 
\begin{tikzpicture}[baseline=(current bounding box.north)]
\pgfplotsset{colormap={rdylgnrev}{rgb=(0.0,0.408,0.216) rgb=(0.0729,0.5423,0.286) rgb=(0.2297,0.6581,0.3457) rgb=(0.4359,0.7567,0.392) rgb=(0.6151,0.8353,0.412) rgb=(0.7653,0.9001,0.4897) rgb=(0.8936,0.955,0.6033) rgb=(1.0,1.0,0.749) rgb=(0.9971,0.9129,0.6033) rgb=(0.9943,0.794,0.4743) rgb=(0.987,0.6456,0.3633) rgb=(0.962,0.4634,0.2797) rgb=(0.8919,0.2904,0.2001) rgb=(0.787,0.1343,0.1519) rgb=(0.647,0.0,0.149)}, colormap={rdylgn}{rgb=(0.647,0.0,0.149) rgb=(0.787,0.1343,0.1519) rgb=(0.8919,0.2904,0.2001) rgb=(0.962,0.4634,0.2797) rgb=(0.987,0.6456,0.3633) rgb=(0.9943,0.794,0.4743) rgb=(0.9971,0.9129,0.6033) rgb=(1.0,1.0,0.749) rgb=(0.8936,0.955,0.6033) rgb=(0.7653,0.9001,0.4897) rgb=(0.6151,0.8353,0.412) rgb=(0.4359,0.7567,0.392) rgb=(0.2297,0.6581,0.3457) rgb=(0.0729,0.5423,0.286) rgb=(0.0,0.408,0.216)}}
\begin{axis}[hide axis, scale only axis, width=0pt, height=1.6cm,
  colormap name=rdylgn, colorbar, point meta min=0.0, point meta max=1.0,
  colorbar style={height=1.6cm, width=0.25cm, ylabel={P(detection)}, ylabel style={font=\scriptsize},
    ytick={0.0,0.25,0.5,0.75,1.0}, yticklabels={0,.25,.5,.75,1}, tick label style={font=\scriptsize},
    yticklabel style={text width=1.7em, align=right}}]
  \addplot[draw=none] coordinates {(0,0)};
\end{axis}
\end{tikzpicture} \\[2pt]
\multicolumn{3}{c}{(a) Probability of Detection} & \\[8pt]
\input{figures/floorplan_occ_50/floorplan_blob_trans} & \input{figures/floorplan_occ_50/floorplan_april_trans} & \input{figures/floorplan_occ_50/floorplan_aruco_trans} & 
\begin{tikzpicture}[baseline=(current bounding box.north)]
\pgfplotsset{colormap={rdylgnrev}{rgb=(0.0,0.408,0.216) rgb=(0.0729,0.5423,0.286) rgb=(0.2297,0.6581,0.3457) rgb=(0.4359,0.7567,0.392) rgb=(0.6151,0.8353,0.412) rgb=(0.7653,0.9001,0.4897) rgb=(0.8936,0.955,0.6033) rgb=(1.0,1.0,0.749) rgb=(0.9971,0.9129,0.6033) rgb=(0.9943,0.794,0.4743) rgb=(0.987,0.6456,0.3633) rgb=(0.962,0.4634,0.2797) rgb=(0.8919,0.2904,0.2001) rgb=(0.787,0.1343,0.1519) rgb=(0.647,0.0,0.149)}, colormap={rdylgn}{rgb=(0.647,0.0,0.149) rgb=(0.787,0.1343,0.1519) rgb=(0.8919,0.2904,0.2001) rgb=(0.962,0.4634,0.2797) rgb=(0.987,0.6456,0.3633) rgb=(0.9943,0.794,0.4743) rgb=(0.9971,0.9129,0.6033) rgb=(1.0,1.0,0.749) rgb=(0.8936,0.955,0.6033) rgb=(0.7653,0.9001,0.4897) rgb=(0.6151,0.8353,0.412) rgb=(0.4359,0.7567,0.392) rgb=(0.2297,0.6581,0.3457) rgb=(0.0729,0.5423,0.286) rgb=(0.0,0.408,0.216)}}
\begin{axis}[hide axis, scale only axis, width=0pt, height=1.6cm,
  colormap name=rdylgnrev, colorbar, point meta min=0.301, point meta max=2.0792,
  colorbar style={height=1.6cm, width=0.25cm, ylabel={trans.~error [mm]}, ylabel style={font=\scriptsize},
    ytick={0.301,0.699,1.0,1.3979,1.7782,2.0792}, yticklabels={2,5,10,25,60,120}, tick label style={font=\scriptsize},
    yticklabel style={text width=1.7em, align=right}}]
  \addplot[draw=none] coordinates {(0,0)};
\end{axis}
\end{tikzpicture} \\[2pt]
\multicolumn{3}{c}{(b) Median Translation Error [mm]} & \\[8pt]
\input{figures/floorplan_occ_50/floorplan_blob_rot} & \input{figures/floorplan_occ_50/floorplan_april_rot} & \input{figures/floorplan_occ_50/floorplan_aruco_rot} & 
\begin{tikzpicture}[baseline=(current bounding box.north)]
\pgfplotsset{colormap={rdylgnrev}{rgb=(0.0,0.408,0.216) rgb=(0.0729,0.5423,0.286) rgb=(0.2297,0.6581,0.3457) rgb=(0.4359,0.7567,0.392) rgb=(0.6151,0.8353,0.412) rgb=(0.7653,0.9001,0.4897) rgb=(0.8936,0.955,0.6033) rgb=(1.0,1.0,0.749) rgb=(0.9971,0.9129,0.6033) rgb=(0.9943,0.794,0.4743) rgb=(0.987,0.6456,0.3633) rgb=(0.962,0.4634,0.2797) rgb=(0.8919,0.2904,0.2001) rgb=(0.787,0.1343,0.1519) rgb=(0.647,0.0,0.149)}, colormap={rdylgn}{rgb=(0.647,0.0,0.149) rgb=(0.787,0.1343,0.1519) rgb=(0.8919,0.2904,0.2001) rgb=(0.962,0.4634,0.2797) rgb=(0.987,0.6456,0.3633) rgb=(0.9943,0.794,0.4743) rgb=(0.9971,0.9129,0.6033) rgb=(1.0,1.0,0.749) rgb=(0.8936,0.955,0.6033) rgb=(0.7653,0.9001,0.4897) rgb=(0.6151,0.8353,0.412) rgb=(0.4359,0.7567,0.392) rgb=(0.2297,0.6581,0.3457) rgb=(0.0729,0.5423,0.286) rgb=(0.0,0.408,0.216)}}
\begin{axis}[hide axis, scale only axis, width=0pt, height=1.6cm,
  colormap name=rdylgnrev, colorbar, point meta min=-0.699, point meta max=1.699,
  colorbar style={height=1.6cm, width=0.25cm, ylabel={rot.~error [$^\circ$]}, ylabel style={font=\scriptsize},
    ytick={-0.699,-0.301,0.0,0.301,0.699,1.1761,1.699}, yticklabels={0.2,0.5,1,2,5,15,50}, tick label style={font=\scriptsize},
    yticklabel style={text width=1.7em, align=right}}]
  \addplot[draw=none] coordinates {(0,0)};
\end{axis}
\end{tikzpicture} \\[2pt]
\multicolumn{3}{c}{(c) Median Rotation Error [$^\circ$]} & \\
\end{tabular}

	\caption{Pose accuracy over the capture floor plan at $50\%$ occlusion, in
		the same layout and encoding as the floor-plan figure of the main paper. Rings are
		tripod distance, wedges are $10^\circ$ rig stations; colour is (a) the
		fraction of the $30$ boards detected, and the per-station median over
		detected boards of (b) translation and (c) rotation error.
		\textbf{Gray wedges are stations with zero detections.} Both tag plans
		are uniformly gray; the BlobBoards plan retains coverage across the
		capture floor at per-station errors in the range it holds unoccluded.}
	\label{fig:supp:occlusion50}
\end{figure*}

\begin{figure*}[h]
	\centering
\providecommand{\fpscale}{1}\renewcommand{\fpscale}{0.42}   
\setlength{\tabcolsep}{3pt}
\centering
\begin{tabular}{@{}ccc l@{}}
\textbf{BlobBoards} & \textbf{AprilTag} & \textbf{ArUco} & \\[2pt]
\input{figures/floorplan_occ_70/floorplan_blob_det} & \input{figures/floorplan_occ_70/floorplan_april_det} & \input{figures/floorplan_occ_70/floorplan_aruco_det} & 
\begin{tikzpicture}[baseline=(current bounding box.north)]
\pgfplotsset{colormap={rdylgnrev}{rgb=(0.0,0.408,0.216) rgb=(0.0729,0.5423,0.286) rgb=(0.2297,0.6581,0.3457) rgb=(0.4359,0.7567,0.392) rgb=(0.6151,0.8353,0.412) rgb=(0.7653,0.9001,0.4897) rgb=(0.8936,0.955,0.6033) rgb=(1.0,1.0,0.749) rgb=(0.9971,0.9129,0.6033) rgb=(0.9943,0.794,0.4743) rgb=(0.987,0.6456,0.3633) rgb=(0.962,0.4634,0.2797) rgb=(0.8919,0.2904,0.2001) rgb=(0.787,0.1343,0.1519) rgb=(0.647,0.0,0.149)}, colormap={rdylgn}{rgb=(0.647,0.0,0.149) rgb=(0.787,0.1343,0.1519) rgb=(0.8919,0.2904,0.2001) rgb=(0.962,0.4634,0.2797) rgb=(0.987,0.6456,0.3633) rgb=(0.9943,0.794,0.4743) rgb=(0.9971,0.9129,0.6033) rgb=(1.0,1.0,0.749) rgb=(0.8936,0.955,0.6033) rgb=(0.7653,0.9001,0.4897) rgb=(0.6151,0.8353,0.412) rgb=(0.4359,0.7567,0.392) rgb=(0.2297,0.6581,0.3457) rgb=(0.0729,0.5423,0.286) rgb=(0.0,0.408,0.216)}}
\begin{axis}[hide axis, scale only axis, width=0pt, height=1.6cm,
  colormap name=rdylgn, colorbar, point meta min=0.0, point meta max=1.0,
  colorbar style={height=1.6cm, width=0.25cm, ylabel={P(detection)}, ylabel style={font=\scriptsize},
    ytick={0.0,0.25,0.5,0.75,1.0}, yticklabels={0,.25,.5,.75,1}, tick label style={font=\scriptsize},
    yticklabel style={text width=1.7em, align=right}}]
  \addplot[draw=none] coordinates {(0,0)};
\end{axis}
\end{tikzpicture} \\[2pt]
\multicolumn{3}{c}{(a) Probability of Detection} & \\[8pt]
\input{figures/floorplan_occ_70/floorplan_blob_trans} & \input{figures/floorplan_occ_70/floorplan_april_trans} & \input{figures/floorplan_occ_70/floorplan_aruco_trans} & 
\begin{tikzpicture}[baseline=(current bounding box.north)]
\pgfplotsset{colormap={rdylgnrev}{rgb=(0.0,0.408,0.216) rgb=(0.0729,0.5423,0.286) rgb=(0.2297,0.6581,0.3457) rgb=(0.4359,0.7567,0.392) rgb=(0.6151,0.8353,0.412) rgb=(0.7653,0.9001,0.4897) rgb=(0.8936,0.955,0.6033) rgb=(1.0,1.0,0.749) rgb=(0.9971,0.9129,0.6033) rgb=(0.9943,0.794,0.4743) rgb=(0.987,0.6456,0.3633) rgb=(0.962,0.4634,0.2797) rgb=(0.8919,0.2904,0.2001) rgb=(0.787,0.1343,0.1519) rgb=(0.647,0.0,0.149)}, colormap={rdylgn}{rgb=(0.647,0.0,0.149) rgb=(0.787,0.1343,0.1519) rgb=(0.8919,0.2904,0.2001) rgb=(0.962,0.4634,0.2797) rgb=(0.987,0.6456,0.3633) rgb=(0.9943,0.794,0.4743) rgb=(0.9971,0.9129,0.6033) rgb=(1.0,1.0,0.749) rgb=(0.8936,0.955,0.6033) rgb=(0.7653,0.9001,0.4897) rgb=(0.6151,0.8353,0.412) rgb=(0.4359,0.7567,0.392) rgb=(0.2297,0.6581,0.3457) rgb=(0.0729,0.5423,0.286) rgb=(0.0,0.408,0.216)}}
\begin{axis}[hide axis, scale only axis, width=0pt, height=1.6cm,
  colormap name=rdylgnrev, colorbar, point meta min=0.301, point meta max=2.0792,
  colorbar style={height=1.6cm, width=0.25cm, ylabel={trans.~error [mm]}, ylabel style={font=\scriptsize},
    ytick={0.301,0.699,1.0,1.3979,1.7782,2.0792}, yticklabels={2,5,10,25,60,120}, tick label style={font=\scriptsize},
    yticklabel style={text width=1.7em, align=right}}]
  \addplot[draw=none] coordinates {(0,0)};
\end{axis}
\end{tikzpicture} \\[2pt]
\multicolumn{3}{c}{(b) Median Translation Error [mm]} & \\[8pt]
\input{figures/floorplan_occ_70/floorplan_blob_rot} & \input{figures/floorplan_occ_70/floorplan_april_rot} & \input{figures/floorplan_occ_70/floorplan_aruco_rot} & 
\begin{tikzpicture}[baseline=(current bounding box.north)]
\pgfplotsset{colormap={rdylgnrev}{rgb=(0.0,0.408,0.216) rgb=(0.0729,0.5423,0.286) rgb=(0.2297,0.6581,0.3457) rgb=(0.4359,0.7567,0.392) rgb=(0.6151,0.8353,0.412) rgb=(0.7653,0.9001,0.4897) rgb=(0.8936,0.955,0.6033) rgb=(1.0,1.0,0.749) rgb=(0.9971,0.9129,0.6033) rgb=(0.9943,0.794,0.4743) rgb=(0.987,0.6456,0.3633) rgb=(0.962,0.4634,0.2797) rgb=(0.8919,0.2904,0.2001) rgb=(0.787,0.1343,0.1519) rgb=(0.647,0.0,0.149)}, colormap={rdylgn}{rgb=(0.647,0.0,0.149) rgb=(0.787,0.1343,0.1519) rgb=(0.8919,0.2904,0.2001) rgb=(0.962,0.4634,0.2797) rgb=(0.987,0.6456,0.3633) rgb=(0.9943,0.794,0.4743) rgb=(0.9971,0.9129,0.6033) rgb=(1.0,1.0,0.749) rgb=(0.8936,0.955,0.6033) rgb=(0.7653,0.9001,0.4897) rgb=(0.6151,0.8353,0.412) rgb=(0.4359,0.7567,0.392) rgb=(0.2297,0.6581,0.3457) rgb=(0.0729,0.5423,0.286) rgb=(0.0,0.408,0.216)}}
\begin{axis}[hide axis, scale only axis, width=0pt, height=1.6cm,
  colormap name=rdylgnrev, colorbar, point meta min=-0.699, point meta max=1.699,
  colorbar style={height=1.6cm, width=0.25cm, ylabel={rot.~error [$^\circ$]}, ylabel style={font=\scriptsize},
    ytick={-0.699,-0.301,0.0,0.301,0.699,1.1761,1.699}, yticklabels={0.2,0.5,1,2,5,15,50}, tick label style={font=\scriptsize},
    yticklabel style={text width=1.7em, align=right}}]
  \addplot[draw=none] coordinates {(0,0)};
\end{axis}
\end{tikzpicture} \\[2pt]
\multicolumn{3}{c}{(c) Median Rotation Error [$^\circ$]} & \\
\end{tabular}

	\caption{Pose accuracy over the capture floor plan at $70\%$ occlusion, in
		the same layout and encoding as the floor-plan figure of the main paper.
		Rings are tripod distance, wedges are $10^\circ$ rig stations; colour is
		(a) the fraction of the $30$ boards detected, and the per-station
		median over detected boards of (b) translation and (c) rotation error.
		\textbf{Gray wedges are stations with zero detections.} The AprilTag
		and ArUco plans are gray everywhere: at this level neither returns a
		pose anywhere on the capture floor. BlobBoards retain coverage over a
		reduced but still substantial set of stations, at per-station errors in
		the same range as the unoccluded case.}
	\label{fig:supp:occlusion70}
\end{figure*}

\end{document}